\documentclass{article}
\usepackage[preprint]{neurips_2026}

\usepackage[utf8]{inputenc} 
\usepackage[T1]{fontenc}    
\usepackage{hyperref}       
\usepackage{url}            
\usepackage{booktabs}       
\usepackage{amsfonts}       
\usepackage{nicefrac}       
\usepackage{microtype}      
\usepackage{xcolor}         
\usepackage{colortbl}       
\usepackage{graphicx}
\usepackage{tikz}
\usetikzlibrary{arrows.meta}
\usepackage{amsmath}
\usepackage{algorithm}
\usepackage{algpseudocode}
\usepackage{cleveref}
\usepackage{tabularx}
\usepackage{multirow}
\usepackage{zref-clever}
\usepackage{wrapfig}
\usepackage{caption}
\usepackage{subcaption}

\zcsetup{
  countertype={
    section=suppsection,
    table=supptable,
    figure=suppfigure,
  },
}

\zcLanguageSetup{english}{
  type = appendix ,
    Name-sg = Appendix ,
    name-sg = Appendix ,
    Name-pl = Appendices ,
    name-pl = Appendices ,
  type = equation ,
    Name-sg = Eq. ,
    name-sg = Eq. ,
    Name-pl = Eq. ,
    name-pl = Eq. ,
  type = section ,
    Name-sg = Section ,
    name-sg = Section ,
    Name-pl = Sections ,
    name-pl = Sections ,
  type = suppsection ,
    Name-sg = Section ,
    name-sg = Section ,
    Name-pl = Sections ,
    name-pl = Sections ,
  type = suppfigure ,
    Name-sg = Figure ,
    name-sg = Figure ,
    Name-pl = Figures ,
    name-pl = Figures ,
  type = supptable ,
    Name-sg = Table ,
    name-sg = Table ,
    Name-pl = Tables ,
    name-pl = Tables ,
}

\newcommand\methoda{ACORN}
\newcommand\methodn{\textbf{A}ctive \textbf{Co}ntinuous-Sco\textbf{r}e Occupancy Modeli\textbf{n}g}

\newcommand\TODO[1]{}

\title{Targeted Review for AI-Assisted Biodiversity Surveys: Active Continuous-Score Occupancy Modeling}

\author{%
    Timm Haucke\\
    MIT\\
    \texttt{haucke@mit.edu}
    \And
    Lauren Harrell\\
    Google Research
    \And
    Justin Kay\\
    MIT
    \AND
    Mary Clapp\\
    The Institute for Bird Populations\\
    \And
    Sara Beery\\
    MIT
}

\begin{document}

\maketitle

\begin{abstract}
\vspace{-0.25cm}
We increasingly use machine learning to label scientific datasets. The models we develop and deploy are improving all the time, but they are not and will likely never be perfect.
Mistakes matter, as errors can propagate into our scientific understanding, particularly when systematically biased.
Very reasonably, scientists thus review \textit{substantial} proportions of ML-generated labels to verify or correct mistakes in pursuit of ensuring their scientific findings are not biased by ML. In this work, we focus on helping scientists allocate review effort efficiently and in alignment with their scientific goals. 
We focus on a specific class of scientists (ecologists) and a specific, widespread, and impactful modeling target (occupancy modeling, which estimates where species are likely to occur, conditioned on environmental factors). 
We introduce \methodn~(\methoda), a method that incorporates ML predictions into occupancy models and strategically selects samples for expert review by expected information gain for downstream ecological analysis. Across camera-trap and bioacoustic datasets, our method recovers ecological conclusions close to those obtained from fully human-labeled data, while requiring substantially fewer expert reviews than non-targeted review policies. Our results suggest that ML-assisted scientific workflows should optimize expert effort for downstream inference, rather than for classifier accuracy alone, especially when human review budget is limited. Our code is available at \url{https://github.com/timmh/acorn}
\end{abstract}

\begin{figure}[H]
  \centering
  \vspace{-0.5cm}
  \input{figures/figure_1.tex}
  \caption{\methoda\ turns imperfect ML classifications into targeted ecological inference. Large-scale camera-trap and bioacoustic surveys produce datasets that are costly to label exhaustively, but naively relying on ML predictions can bias occupancy estimates and covariate conclusions. \methoda\ combines continuous classifier scores with a hierarchical occupancy model, actively prioritizes expert review using goal-directed Bayesian experimental design, and evaluates progress using ecological quantities that matter for downstream decision-making: occurrence probabilities, site rankings, and occupancy and detection effects.}
  \vspace{-0.5cm}
  \label{fig:graphical_abstract}
\end{figure}

\section{Introduction}\label{sec:introduction}

Machine learning (ML) is rapidly changing how ecological data is processed \citep{tuia2022perspectives}. Data from camera traps and passive acoustic sensors, which passively collect images and audio \textit{in situ}, are increasingly collected at such massive scales that they are often infeasible to review manually. ML models can classify species per example (e.g. image or audio window) far quicker than human experts \citep{norouzzadeh2018automatically,tabak2019machine,kahl2021birdnet}. However, in many ecological studies, the final objective is not per-example classification (e.g. identifying every species in every image of the dataset). Rather, the goal is often to use per-example classifications for population-level inference, such as estimating whether a species is present at a site (\emph{occupancy}), how its detectability varies with survey conditions, and relating environmental factors (covariates) to the probability of a species occurring in a given place \citep{yoccoz2001monitoring,nichols2006monitoring}.

Population-level statistical ecological models typically assume the per-example classifications they are built on are (by construction) correct. However, when we move to ML-generated classifications, this assumption breaks. Small error rates in per-example classification, i.e. even a single missed detection or false positive, can have substantial effects on downstream inference depending on the place, time, and type of mistake, biasing estimates and distorting estimated covariate relationships \citep{mackenzie2002estimating,royle2006generalized,santoro2025essential}. As a result, many ecologists are reasonably reluctant to rely on ML-generated observations in ecological workflows~\cite{velez2023evaluation}. In practice, classifiers are frequently used to filter obvious negatives or otherwise de-prioritize uninformative samples, while experts continue to manually review large fractions of data \citep{santoro2025essential}. This review burden is expensive and slow, limiting the extent to which data collected from large-scale sensor networks translates into timely ecological insights or conservation actions \citep{norouzzadeh2021deep,gibb2019emerging}.

Instead, we seek to use ML outputs directly in the downstream ecological model while explicitly accounting for classifier error. As an illustrative example we focus on occupancy models, hierarchical latent variable models that disentangle the probability of species presence from the probability of observing it.
Because occupancy models facilitate unbiased inference from noisy observations, their use is widespread in ecology and conservation. They are explicitly recognized in IUCN Red List guidance as an appropriate tool for quantitative extinction-risk analysis \citep{iucn_redlist_guidelines}, and have been applied to invasive species monitoring \cite{bajcz2024within}, identifying high-priority conservation areas \cite{de2009using}, and used to inform conservation management \cite{fuller2016management}. However, our approach is adaptable to, and could be similarly impactful for, other population-level inference tasks built from ML-generated classifications.

Recent work has shown that it is possible to combine human reviews and ML predictions to estimate occupancy more efficiently than with human reviews alone \citep{rhinehart2022continuous}. 
But the question remains: \textit{which datapoints are most useful to review?}
Building heuristics for prioritizing which reviews should be collected is extremely difficult, since the information each reviewed example provides depends on where and when it was captured, the ecological context, and how redundant it is with respect to previously acquired reviews. Bayesian experimental design~\cite{lindley1956measure,chaloner1995bayesian} provides a natural information-theoretic framework to prioritize the collection of new observations (in our case, human reviews) that maximize information gained about a quantity of interest.

We propose \methodn\ (\methoda), a method that combines noisy machine learning classifications with targeted review to maximize information gain towards a user-specified ecological inference target, based on Bayesian experimental design. 
Rather than targeting review to optimize classifier performance, as you would in active learning, our review acquisition policy is constructed to align with the posterior quantities that matter for ecological analysis, such as occupancy parameters, detection parameters, and derived occurrence estimates.
We additionally propose a stopping criterion to determine when additional expert effort is no longer worthwhile, motivated by practical deployment needs.
We systematically benchmark our method on two representative ecological datasets capturing observations of species across deployed networks of static sensors, one capturing species observations through images and one through bioacoustic recordings. We compare methods that only use verified reviews for inference and methods which additionally make use of ML scores on unverified data, and across both inference methods compare five different review policies. Evaluated across four metrics that measure recovery of ecological conclusions, we find that our method, \methoda\, which uses ML scores across the data to inform a targeted information gain review policy, generally converges more quickly towards estimates consistent with fully labeled datasets.
Overall, this work emphasizes the importance of developing systems that enable scientists to work \textit{with} imperfect models to improve \textit{targeted} scientific understanding.

\section{Related work}\label{sec:related_work}
\noindent\textbf{ML-based species identification and occupancy modeling.}
ML models can label camera-trap and acoustic biodiversity data at scale, but deployment remains limited by frequent errors related to geographic or visual shifts, rare species, and poorly-calibrated predictions \citep{norouzzadeh2018automatically,tabak2019machine,willi2019identifying,beery2019iwildcam,beery2018recognition,kahl2021birdnet,perch_v1}.
ML scores are often used to speed up human labeling via thresholding, filtering, or active learning \citep{norouzzadeh2018automatically,whytock2021robust,norouzzadeh2021deep,bothmann2023automated,kholghi2018active,kath2024leveraging,miao2021iterative,gordon2023}. However, nearly all such pipelines are still classifier-centric: they choose thresholds or queried examples to improve classification performance, domain adaptation, or annotation efficiency, rather than to directly reduce uncertainty in a downstream ecological posterior.

\noindent\textbf{Occupancy modeling with imperfect detection and uncertain labels.}
Occupancy models are a natural target when moving beyond per-example classification toward population-level inference. They separate latent site occupancy from imperfect detection, 
allowing a species to be present at a site even if it is not detected during one or more repeated surveys, or ``replicates'', and they allow occupancy and detection probabilities to depend on ecological covariates \citep{mackenzie2002estimating}. Classical occupancy models account for false negatives through imperfect detection, but ML-derived observations also introduce false positives. Ignoring these additional error processes can materially bias inference \citep{royle2006generalized,miller2011improving,ferguson2015occupancy,chambert2018two,wright2020modelling}, though occupancy inference has been shown to be robust to classifier errors in some settings~\citep{bevan2025deep,thornton2026,katsis2025comparison}. 
Recently, a joint continuous-score model was proposed which directly models ML score distributions conditional on latent detection states, optionally anchored by human labels \cite{rhinehart2022continuous}. This retains more information than thresholding and provides a probabilistic interface between ML outputs and ecological inference. However, that framework does not address the practical allocation problem created by limited expert review: which samples should be prioritized by experts, and when can review stop? Prior design work for occupancy studies has optimized field visits and survey schedules \citep{mackenzie2002estimating,guillera2014two}. In contrast, we treat expert verification itself as the design problem: given an existing set of ML-scored observations, which sample should be reviewed next?

\noindent\textbf{Prediction-assisted inference under scarce labels.}
Methods that use ML predictions to inform downstream statistical inference with few verified labels include post-prediction inference~\citep{postpi}, semi-supervised and surrogate-outcome methods \citep{gronsbell2018semi}, and prediction-powered inference (PPI), which combines a small labeled sample with many ML predictions to obtain tighter confidence intervals for low-dimensional population parameters \citep{angelopoulos2023prediction}. Standard PPI formulations are typically fixed-design: labeled and unlabeled samples are representative draws from the target population, and labels serve as a correction or calibration sample. Follow-up work extends this paradigm to actively collected labels \citep{zrnicactive}. Our approach is complementary but distinct: labels are acquired intentionally non-representatively to reduce uncertainty in an entire ecological posterior.

\noindent\textbf{Bayesian experimental design, Bayesian optimal design, and Bayesian active learning.}
Bayesian experimental design underpins our decision-theoretic approach for active review, prioritizing  experiments, measurements, or labels by maximizing expected utility under the current posterior and predictive distribution \citep{lindley1956measure,chaloner1995bayesian,oaksford2003optimal,ryan2016review,rainforth2024modern}. In the canonical information-theoretic formulation, the utility is expected information gain, or expected posterior entropy reduction. Classical works distinguish parameter-, prediction-, and decision-focused utilities, which matters when the target is not classifier accuracy but an ecological summary \citep{chaloner1995bayesian}. Bayesian active learning methods such as BALD select labels informative about the model posterior \citep{houlsby2011bayesian}. Bayesian experimental design underlies related work in active testing and model selection \citep{kossen2022active,kay2025consensus}. We adapt this view to biodiversity surveys: the experiment is asking an expert to verify one ML-scored replicate, the outcome is a binary detection / non-detection label, and the utility targets ecological conclusions rather than classifier performance.

\section{Methods}\label{sec:methods}
\methodn\ treats expert review prioritization for ML-supported species occupancy inference as an iterative Bayesian experimental design problem. Two main design components of \methoda\ are the ecological model (\zcref{sec:methods:models}) and the review policy (\zcref{sec:methods:policies}).

\subsection{The ecological model}\label{sec:methods:models}

The classic Bernoulli occupancy model \citep{mackenzie2002estimating} models latent occupancy of site $i \in S$ as
\begin{equation}
    z_i \sim \mathrm{Bernoulli}(\psi_i)
\end{equation}
where $\psi_i$ is the probability of occupancy and is modeled using logistic regression:
\begin{equation}
    \log\left(\frac{\psi_{i}}{1-\psi_{i}} \right) = \beta_0 + X_i^\top \beta
\end{equation}

To account for the fact that species can be present even if unobserved, we further model a detection probability $p_{ij}$ for replicate $j \in \{1, ...,J_i\}$:
\begin{equation}
    \log\left(\frac{p_{ij}}{1-p_{ij}} \right) = \alpha_0 + W_{ij}^{\top} \alpha
\end{equation}

Occupancy covariates $X_i$ vary per site and include environmental variables like average temperature or elevation. Detection covariates $W_{ij}$ vary per site and per replicate and include factors that influence how easy a species is to detect, such as time of day. The full set of covariates used across acoustic and camera trap models can be found in~\zcref{sec:appendix:covariates}.

Expert-verified binary labels $f_{ij}$ indicate whether the focal species was detected at site $i$ and replicate $j$. These detections are then modeled conditional on the occupancy status:
\begin{equation}
    f_{ij} | z_i \sim \mathrm{Bernoulli}(z_i p_{ij})
\end{equation}

\noindent\textbf{Our decoupled continuous-score occupancy model.}

We now introduce our decoupled continuous-score occupancy model, which builds upon the classic Bernoulli occupancy model to incorporate expert-reviewed observations and upon the joint continuous-score occupancy model \cite{rhinehart2022continuous} to utilize noisy ML classifier scores. To do so, we specify a score model based on two class-conditional Normal densities,
\begin{equation}
s_{ij} \mid f_{ij}=0 \sim \mathcal{N}(\mu_0, \sigma_0^2),
\qquad
s_{ij} \mid f_{ij}=1 \sim \mathcal{N}(\mu_1, \sigma_1^2),
\qquad \mu_1 > \mu_0,
\end{equation}
where the lower component corresponds to observations without the focal species and the upper component to replicates containing the focal species.

The original joint formulation \cite{rhinehart2022continuous} estimates the score parameters $(\mu_0,\sigma_0,\mu_1,\sigma_1)$ and the ecological parameters in one posterior. This is statistically appealing when the Normal score model is well specified, because occupancy structure and score separation can inform one another. In our data, however, the score distribution is often much better identified by the marginal classifier-score distribution and the reviewed labels than by the ecological process itself. If the score mixture is misspecified, joint estimation can therefore create a problematic feedback loop: the occupancy and detection model can pull the score components toward a configuration that explains the spatial pattern of scores, while the distorted score components then feed back into occupancy inference.

We therefore use a ``decoupled'' version of the Normal score model. At each review round, we first fit the two-component score model to the finite classifier scores and the currently reviewed labels. Unreviewed scores contribute through the marginal mixture likelihood, while reviewed scores are assigned to the component indicated by the expert label. This produces a posterior $q^{(t)}(\eta)$ over the score-calibration parameters $\eta=(\pi,\mu_0,\sigma_0,\mu_1,\sigma_1)$, where $\pi$ is the marginal prevalence of true detections in the score distribution. Crucially, this posterior is conditioned only on scores and reviewed labels, not on the occupancy likelihood. The ecological model subsequently receives the resulting score evidence, but it cannot update the calibration parameters. In this sense, information flows from the score calibration into the occupancy model, while there is no feedback from occupancy into calibration.

For each score, the calibrated evidence is summarized as a log Bayes factor
\begin{equation}
\ell_{ij}
=
\log \mathbb{E}_{\eta\sim q^{(t)}}\!\left[p(s_{ij} \mid f_{ij}=1,\eta)\right]
-
\log \mathbb{E}_{\eta\sim q^{(t)}}\!\left[p(s_{ij} \mid f_{ij}=0,\eta)\right].
\end{equation}
A value $\ell_{ij}=0$ is neutral evidence, $\ell_{ij}>0$ means the score is more typical of a true detection, and $\ell_{ij}<0$ means the score is more typical of a non-detection. For acquisition functions, we also retain a small ensemble of calibration draws $\ell_{ij}^{(r)}$ so that uncertainty in score calibration contributes to uncertainty about candidate labels.

Conditional on an occupied site, an unreviewed score adds directly to the log odds that replicate $j$ contains the focal species:
\begin{equation}
\log\left(
\frac{\Pr(f_{ij}=1 \mid z_i=1, s_{ij})}{1-\Pr(f_{ij}=1 \mid z_i=1, s_{ij})}
\right)
=
\log\left(\frac{p_{ij}}{1-p_{ij}}\right) + \ell_{ij}.
\end{equation}
Equivalently, an unreviewed score contributes the replicate-level occupancy evidence $(1-p_{ij}) + p_{ij}\exp(\ell_{ij})$ after the common negative-score density has been absorbed into the normalizing constant. In other words, a high score is not treated as ultimate proof that the site is occupied, rather, it is treated as evidence that this particular replicate may contain the species, which in turn makes site occupancy more plausible. When a replicate is reviewed, we use the expert label directly in the same Bernoulli detection model as above. Intuitively, this replaces the probabilistic evidence of the score with definite evidence of the reviewed human label. The reviewed score remains useful for calibrating future scores, but once the label is available, the score is not also counted as separate evidence about whether the site is occupied.

\subsection{The review policy}\label{sec:methods:policies}

We interpret choosing the next replicate to review as a Bayesian experimental design problem. Let $a$ denote a candidate review action and let $\theta$ denote the set of model parameters and latent variables. A natural utility is the expected information gain
\begin{equation}
\mathrm{EIG}(a)
=
H\!\left[p(\theta \mid \mathcal{D}^{(t)})\right]
-
\mathbb{E}_{f_a \sim p(f_a \mid \mathcal{D}^{(t)})}
\!\left[
H\!\left[p(\theta \mid \mathcal{D}^{(t)}, f_a)\right]
\right],
\end{equation}
where $\mathcal{D}^{(t)}$ denotes all information available after $t$ review rounds. This criterion asks which review is expected to reduce posterior uncertainty across the entire posterior the most.

However, ecological analyses often care about a more specific set of quantities rather than the full posterior. \textbf{We therefore propose a targeted expected information gain criterion (which we call Target EIG)} that focuses on the specific set of quantities we care about. We define a target vector $\phi=g(\theta)$ containing the site-level occupancy probabilities and the occupancy and detection regression coefficients, and score a candidate by
\begin{equation}
\mathrm{TargetEIG}(a)
=
H\!\left[p(\phi \mid \mathcal{D}^{(t)})\right]
-
\mathbb{E}_{f_a \sim p(f_a \mid \mathcal{D}^{(t)})}
\!\left[
H\!\left[p(\phi \mid \mathcal{D}^{(t)}, f_a)\right]
\right].
\end{equation}
Thus, Target EIG prioritizes reviews expected to reduce uncertainty in the ecological quantities evaluated in \zcref{sec:experiments:evaluation}. We approximate it with posterior draws by estimating the entropy of $\phi$ with a Gaussian approximation. For each candidate review, the hypothetical positive and negative outcomes reweight the existing posterior draws according to the candidate-specific posterior probabilities $\Pr(f_a=1\mid \theta,\mathcal{D}^{(t)})$ and $\Pr(f_a=0\mid \theta,\mathcal{D}^{(t)})$, avoiding a full model refit for every possible label. Because $\phi$ includes many site-level occupancy probabilities, we compute this entropy after standardizing the target summaries and projecting them to a low-dimensional principal-component representation that preserves most posterior variation. To anchor the positive score distribution, Target EIG and BALD initially select max-score batches, switching to their respective acquisition criteria once at least one positive label has been verified. The resulting policy is greedy, batched and uses approximate utilities. As a consequence, it has no guarantee of global optimality over a finite review budget.

\section{Experiments}\label{sec:experiments}

\subsection{Data}\label{sec:experiments:data}
We evaluate on one dataset of passive-acoustic recordings and one camera-trap dataset, using the same downstream review loop in both settings. \textbf{Acoustic:} We use the acoustic forest soundscape dataset \citep{enaa} together with Perch v2 \citep{perch_v2} outputs.
\textbf{Camera trap:} We use the iWildCam 2022 camera-trap dataset \citep{beery2021iwildcam} together with SpeciesNet \citep{speciesnet} outputs.
Each camera trap is a site, and each day of collected data represents a replicate.
Both datasets are represented as single-species, single-season occupancy problems. We fit individual models across the 50 most prevalent bird species in the acoustic data and across the 25 most prevalent species in the camera trap data. This eliminates a long tail of extremely rare species that are difficult to model, even with complete labels, and thus difficult to evaluate. The acoustic and camera-trap experiments otherwise differ only in how data is filtered, how replicate-level scores and labels are constructed, and which covariates are used. More details can be found in \zcref{sec:appendix:datasets}. Considerations and practical guidance for applying \methoda\ to other datasets is provided in \zcref{sec:appendix:practical}.

\subsection{Baselines}\label{sec:experiments:baselines}

\paragraph{Ecological models.}
We compare our proposed decoupled continuous-score occupancy model to two baseline models: the original Bernoulli model fit only using reviewed labels and no ML classifier scores (\zcref{sec:methods:models}), and the original joint continuous-score model \cite{rhinehart2022continuous}.

\paragraph{Review policies.}
We compare our proposed Target EIG policy to another information-theoretic review policy: BALD \citep{houlsby2011bayesian}, a computationally simpler equivalent to the expected information gain (\zcref{sec:methods:policies}) that does not require estimating posterior distributions for each candidate review:
\begin{equation}
\mathrm{BALD}(a)
=
I(f_a ; \theta \mid \mathcal{D}^{(t)})
=
H\!\left[p(f_a \mid \mathcal{D}^{(t)})\right]
-
\mathbb{E}_{\theta \sim p(\theta \mid \mathcal{D}^{(t)})}
\!\left[
H\!\left[p(f_a \mid \theta, \mathcal{D}^{(t)})\right]
\right],
\end{equation}
where $H$ denotes Shannon entropy. It favors observations whose reviewed label is both uncertain under the full current posterior and diagnostically useful for reducing that uncertainty.

We furthermore compare with three simpler baseline policies. \emph{Random} review samples uniformly across reviewable replicates without replacement. \emph{Max-score} selects the reviewable replicates with maximum classifier scores and is a heuristic commonly deployed in ecological practice (since positive observations are frequently rare and thus particularly valuable). \emph{Posterior predictive uncertainty} uses

\begin{equation}
\mathrm{PPU}(a) = H\!\left[p(f_a \mid \mathcal{D}^{(t)})\right]
\end{equation}

and selects the $m$ reviewable replicates with largest utility. This captures ambiguity in the next reviewed label without explicitly asking whether that ambiguity is informative about the posterior.

\subsection{Evaluation}\label{sec:experiments:evaluation}

Our comparison target for models fit on partially reviewed data is a Bernoulli occupancy model (\zcref{sec:methods:models}) fit on all ground-truth dataset labels and without ML scores (the \textit{oracle}). In our evaluation, we focus on metrics that directly evaluate the agreement in ecological conclusions with the oracle. We denote posteriors after $t$ reviews using superscripts $(t)$ and the corresponding oracle posteriors using superscripts of $O$. For site $i$, let $\bar{\psi}_i^{(t)}$ and $\bar{\psi}_i^O$ be the posterior mean occupancy probabilities, let $S$ be the set of sites, and write $\bar{\psi}^{(t)}_S$ and $\bar{\psi}^{O}_S$ for the corresponding vectors over $S$. For a coefficient $\omega$, define the 90\% credible-interval conclusion using the quantile function $q$:
\begin{equation}\label{eq:sign}
c(\omega)=
\begin{cases}
+, & q_{0.05}(\omega) > 0,\\
-, & q_{0.95}(\omega) < 0,\\
0, & \text{otherwise,}
\end{cases}
\end{equation}
where $+$, $-$, and $0$ denote positive, negative, and uncertain effects, respectively.

Site-level recovery is measured both on the probability scale and by the induced ranking of sites:
\begin{equation}\label{eq:occ}
A_\psi = 1-\frac{\lVert\bar{\psi}^{(t)}-\bar{\psi}^{O}\rVert_1}{|S|},
\qquad
\rho_\psi = \rho\left(R\left[\bar{\psi}^{(t)}\right],R\left[\bar{\psi}^{O}\right]\right)
\end{equation}

The occupancy agreement $A_\psi$ compares posterior mean occupancy probabilities directly to the oracle and thus assesses overall prevalence, while the site-rank Spearman correlation $\rho_\psi$ asks whether the current posterior ranks high- and low-occupancy sites in the same order as the oracle and thus has implications for prioritizing sites for conservation and restoration. For covariate conclusions, we compute the fraction of coefficients with the same positive, negative, or uncertain conclusion as the oracle, separately for occupancy and detection covariates:
\begin{equation}\label{eq:cov}
A_\beta = \mathrm{mean}\left(\mathbf{1}\!\left\{c(\beta^{(t)})=c(\beta^{O})\right\}\right),\qquad
A_\alpha = \mathrm{mean}\left(\mathbf{1}\!\left\{c(\alpha^{(t)})=c(\alpha^{O})\right\}\right)
\end{equation}

We furthermore denote the average of $A_\psi,\ \rho_\psi,\ A_\beta,\ A_\alpha$ as $\overline{A}$.

\subsection{Stopping criteria}\label{sec:experiments:stopping_criteria}
We perform a post-hoc analysis of three stopping criteria on the Target EIG review trajectories. For a candidate stopping point $t$, we compute regret separately for the four oracle-agreement metrics (\zcref{sec:experiments:evaluation}). The plotted regret is the mean difference between the best value achieved by that trajectory and the value at $t$, averaged over these four metrics.

We evaluate each criterion at different operating points by sweeping over a set of criterion-specific thresholds. Let $\bar{R}^{(t)}$ be the MCMC $\hat{R}$ diagnostic at step $t$, $G^{(t)}$ be the maximum acquisition score among reviewable samples, and $\Delta^{(t)}$ be the mean posterior distance from the previous review step. The \emph{MCMC diagnostic} stops at the first step with $\bar{R}^{(t)} \le \tau_R$. The \emph{expected-gain} criterion stops after two consecutive steps with $G^{(t)}/G_0 \le \tau_G$. The \emph{posterior-shift} criterion stops after two consecutive steps with $\Delta^{(t)}/\Delta_0 \le \tau_\Delta$.

\subsection{Review loop}\label{sec:experiments:review_loop}
At each review step $t$, we select $m$ replicates for review according to the review policy. We use $m=5$ for the acoustic and $m=25$ for the camera trap data, balancing computational cost and update frequency (\zcref{sec:appendix:batch_size}). We reveal the true labels corresponding to the $m$ replicates to simulate review and refit the occupancy model with this added information, yielding a new posterior at $t+m$. We run each review trajectory with three random seeds per species and configuration.

\section{Results and discussion}\label{sec:results}

\subsection{The positive impact of incorporating ML scores}\label{sec:results:model_families}

\begin{figure}
    \vspace{-0.5cm}
    \centering
     \includegraphics[width=1\linewidth]{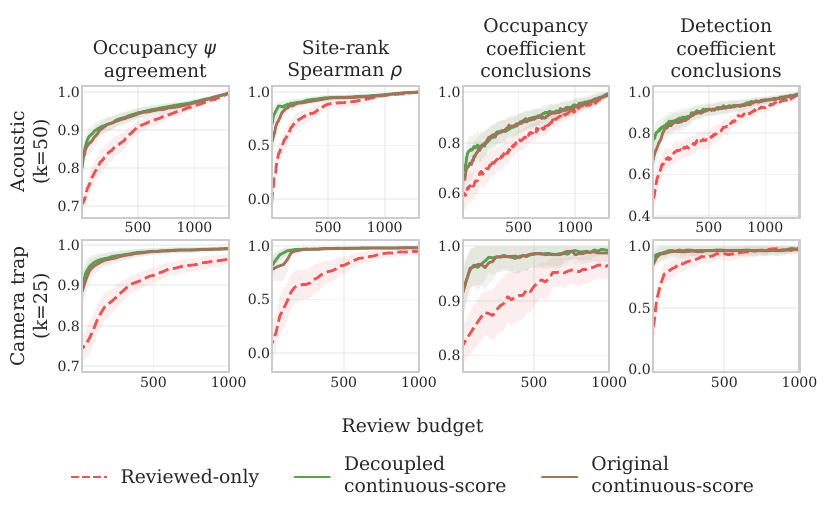}
     \vspace{-0.5cm}
    \caption{\textbf{Using ML scores as noisy evidence enables continuous-score models to start off closer to the ecological conclusions made by the oracle model, and converge more quickly.} The x-axis shows the total number of reviews, while the y-axis shows agreement with the oracle averaged over review policies and species (higher is better). Shaded bands show pointwise 95\% normal-approximation confidence intervals (CIs) across species, after averaging seeds and then shared review policies (BALD and Target EIG) within each species. Acoustic reviews go up to 100\% of available data, while camera trap reviews are capped at 1,000 reviews.}
\label{fig:coefficient_conclusion_agreement_family_average}
\end{figure}

The decoupled continuous-score model (which benefits from information contained in classifier scores) begins much closer to the oracle posterior on average across species than the reviewed-only Bernoulli model (\zcref{fig:coefficient_conclusion_agreement_family_average}). Even before reviews, mean agreement across the four ecological-conclusion metrics (\zcref{sec:experiments:evaluation}) increases from 0.441 to 0.735 for acoustic and from 0.422 to 0.886 for camera traps when comparing the reviewed-only model to the decoupled continuous-score model. This gain enables our targeted review to start from a stronger posterior. Interestingly, early oracle agreement is only moderately associated with classifier quality, suggesting that even imperfect predictions can be useful early on. After review, oracle agreement is consistently high irrespective of classifier quality (\zcref{sec:appendix:classifier}).

How scores enter the ecological model also matters. Within the continuous-score family, the original joint formulation \cite{rhinehart2022continuous}  benefits from classifier scores, but is less robust when the score model is misspecified (\zcref{sec:methods:models}).
The underlying cause for this misspecification is score-distribution heterogeneity: nominal non-detections at occupied sites had higher scores than non-detections at likely unoccupied sites (sites without detections) for the majority of species. Factors that contribute this heterogeneity include label errors (\zcref{fig:mislabeled_examples}) and potential classifier bias. Under joint estimation, this misspecification can create a feedback loop where the occupancy and detection model pull the shared score components toward a configuration that explains spatial score patterns, and the distorted score components then feed back into occupancy inference. Decoupling cuts that feedback by fitting score calibration outside the occupancy posterior and particularly improves oracle agreement at low review counts (\zcref{fig:coefficient_conclusion_agreement_family_average}). This improvement is particularly strong among acoustic species in the top quartile of heterogeneity: agreement improved by 0.115 at zero reviews and 0.059 at 100 reviews, compared with 0.059 and 0.025 among the remaining species. We discuss this phenomenon in detail in \zcref{sec:appendix:score_heterogeneity}.

\subsection{Targeting information gain in review increases efficiency}\label{sec:results:review_strategies}

\begin{table}[t]
\caption{\textbf{ACORN reaches oracle agreement thresholds with few reviews.} Median number of reviews required per species to reach dataset-specific agreement thresholds. Results of the best-performing method(s) are bolded. Thresholds are shown beside each metric name and are higher for camera-trap datasets, where lower thresholds are frequently reached at zero reviews by continuous-score models. The speedup column reports the decoupled continuous-score max-score median review count divided by the \methoda\ median review count.}
\label{tab:agreement_table}
\centering
\begingroup
\small
\setlength{\tabcolsep}{3pt}
\begin{tabular}{@{}lrrrr|rrrrr@{}}
\toprule
 & \multicolumn{4}{c}{\shortstack{Reviewed-only\\Bernoulli}} & \multicolumn{3}{c}{\shortstack{Decoupled\\continuous-score}} &  &  \\
\cmidrule(lr){2-5}\cmidrule(lr){6-8}
Metric & Random & \shortstack{Max-score} & BALD & \shortstack{Target\\EIG} & Random & \shortstack{Max-score} & BALD & \shortstack{\methoda\\(ours)} & \shortstack{Speedup vs.\\max-score} \\
\midrule
\multicolumn{10}{l}{\textbf{Acoustic}} \\
$A_\psi$ $\ge 0.90$ & 340 & 125 & 295 & 205 & 305 & 130 & 65 & \textbf{35} & 3.7$\times$ \\
$\rho_\psi$ $\ge 0.90$ & 365 & 120 & 210 & 175 & 100 & 25 & 20 & \textbf{15} & 1.7$\times$ \\
$A_\beta = 1$ & 565 & 180 & 325 & 265 & 400 & 175 & 105 & \textbf{60} & 2.9$\times$ \\
$A_\alpha = 1$ & 525 & 695 & 370 & 615 & 165 & 35 & \textbf{25} & 45 & 0.8$\times$ \\
\midrule
\multicolumn{10}{l}{\textbf{Camera trap}} \\
$A_\psi$ $\ge 0.95$ & 825 & 750 & 425 & 325 & 375 & 125 & 75 & \textbf{50} & 2.5$\times$ \\
$\rho_\psi$ $\ge 0.95$ & 600 & 400 & 250 & 300 & 200 & 75 & \textbf{50} & \textbf{50} & 1.5$\times$ \\
$A_\beta = 1$ & 275 & 25 & 225 & 175 & \textbf{0} & \textbf{0} & \textbf{0} & \textbf{0} & -- \\
$A_\alpha = 1$ & 100 & 850 & 50 & 100 & \textbf{0} & \textbf{0} & \textbf{0} & \textbf{0} & -- \\
\bottomrule
\end{tabular}

\endgroup
\end{table}

\begin{figure}
    \vspace{-0.5cm}
    \centering
    \includegraphics[width=1\linewidth]{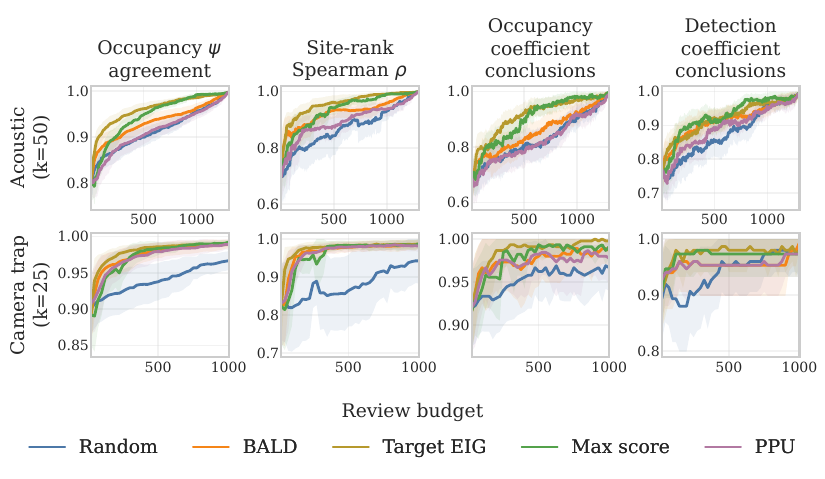}
    \vspace{-0.5cm}
    \caption{\textbf{Target EIG accelerates oracle agreement, especially at low review budgets.} Agreement in ecological conclusions over the number of reviews across review policies. We measure agreement in ecological conclusions by comparing the overall occupancy probability  (\zcref{eq:occ}), the agreement in ranking of individual sites by occupancy probability (\zcref{eq:occ}), the significance and sign of occupancy coefficients (\zcref{eq:cov}), and significance and sign of detection coefficients (\zcref{eq:cov}). Results are averaged over all species and shaded bands show pointwise 95\% normal-approximation CIs across species after averaging seeds within species. Colors indicate the review policy (PPU = posterior predictive uncertainty).}
    \label{fig:coefficient_conclusion_agreement}
\end{figure}

When combined with the decoupled continuous-score occupancy model, Target EIG results in the quickest overall convergence to the fully reviewed oracle, particularly in low- and medium-budget regimes (\zcref{fig:coefficient_conclusion_agreement}). Its advantage over BALD comes from target alignment: BALD favors hard-to-predict labels, but difficult replicates are not necessarily useful for ecological inference. They may be already-resolved, redundant, or unlikely to change the posterior. Target EIG instead scores expected entropy reduction in ecological target quantities, so it can choose lower-uncertainty labels if they better resolve site states, detection processes, or covariate conclusions. Acquisition diagnostics support this interpretation: Target EIG selections are neither simply the highest-BALD nor highest-score reviews, often choosing lower-entropy points than BALD while avoiding many nearly certain high-score detections. Interestingly, focusing EIG on a single task does not consistently improve over the combined objective (\zcref{sec:appendix:task_characterization}).

Despite its simplicity, max-score is a surprisingly strong baseline. Confirmed detections are rare in our data, and positive reviews can anchor site occupancy, inform detection probability, and supply signal for covariate effects. Because classifier scores already rank likely detections well, max-score prioritizes many labels that occupancy models need. This effectiveness of positive-focused review is consistent with prior findings in active learning of classification models for conservation \cite{gordon2023}. The weakness of max-score review is redundancy: the highest-score labels are often predictable from the continuous-score model and can repeat information across visits or sites. Thus Target EIG leads at small budgets, while max-score catches up, and sometimes overtakes Target EIG on particular thresholds, once positive-label accumulation outweighs redundancy.

Overall, our results imply substantial expert time savings. For each metric in \zcref{tab:agreement_table}, \methoda’s median review requirement is at most 60 acoustic or 50 camera-trap reviews. The corresponding maxima for max-score review are 175 and 125 reviews, respectively. For acoustic data, where each replicate is a ten-minute recording, \methoda\ reduces review time from 217.0 hours for all 1,302 recordings to 10.0 hours, saving 207.0 expert-hours per species. Relative to max-score, it saves 115 reviews, or 19.2 hours. This assumes a conservative 1 minute of annotation effort for 1 minute of audio \cite{shaw2022refining}. For camera traps, the iWildCam location-day replicates used in our sweep contain 35.4 images on average. Assuming one expert-hour per 500 images \cite{fennell2022use}, \methoda\ reduces expected review time from 125.4 hours to 3.5 hours, saving 121.9 expert-hours per species. Relative to max-score, it saves 75 reviews, or 5.3 hours per species. Reviewing only the highest-scoring example can reduce effort further, with near-equal results on camera traps but substantially different results on acoustics (\zcref{sec:appendix:review_granularity}).

\subsection{Our stopping criterion balances cost and value}

\begin{figure}
    \centering
    \includegraphics[width=\linewidth]{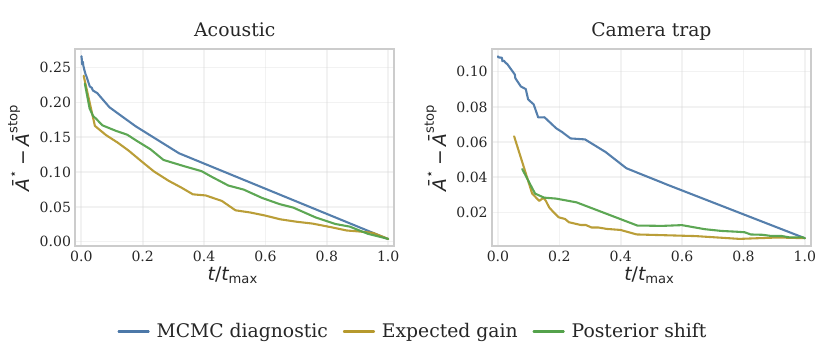}
    \caption{\textbf{Expected gain and posterior shift are better stopping criteria than MCMC diagnostics.} Post-hoc stopping-rule tradeoff for the continuous-score Target EIG trajectories. Each line shows, for one stopping criterion and one dataset, the mean fraction of the full review budget used versus agreement regret relative to the best observed step on that same trajectory, averaged across the four agreement metrics used in \zcref{fig:coefficient_conclusion_agreement}. Criteria farther down and to the left are preferable.}
    \label{fig:budget_regret_tradeoff}
\end{figure}

To be useful in practice, \methoda\ needs to know when to stop reviewing. Crucially, we cannot rely on the oracle model, since that requires datasets that are labeled upfront, thereby defeating the purpose of active review. We therefore evaluate three intuitive stopping criteria based on how well they recover most of the benefit of a full review process (\zcref{fig:budget_regret_tradeoff}). The tradeoff curves show that low $\hat{R}$ is best interpreted as a screen for whether a fitted model is usable, not as evidence that additional labels have stopped changing the ecological conclusions. The expected-gain and posterior-shift rules more directly track whether review remains scientifically useful: the former asks whether any unreviewed sample still has substantial acquisition value, while the latter asks whether successive review steps continue to move the ecological posterior.

These results suggest that stopping should be treated as a decision problem rather than as a convergence diagnostic. Conservative thresholds spend more of the review budget but reduce agreement regret, while aggressive thresholds save review effort at the cost of stopping before coefficient and detection conclusions have stabilized. Without an oracle, practitioners can set tolerances for changes in occupancy probabilities, site rankings, and coefficient conclusions across review batches, then choose expected-gain or posterior-shift thresholds accordingly, allowing species-specific stopping counts.

\section{Conclusion}\label{sec:conclusion}
We propose \methodn\ (\methoda), a method that combines a novel occupancy model formulation that robustly incorporates ML classifier scores with a novel information-theoretic active review policy that prioritizes reviews by estimated information gain about ecological quantities of interest. Across a systematic comparison to alternative combinations of ecological models and review policies on datasets covering two ecological data modalities (camera trap images and acoustic recordings), \methoda\ both provides a much stronger initial starting point and better prioritizes expert reviews, resulting in models that recover oracle-like ecological posteriors with substantially less manual review than unguided strategies.

\noindent\textbf{Limitations and future work.}
Future work could extend targeted review beyond occupancy to abundance estimation with N-mixture models \cite{royle2004n} or distance sampling \cite{buckland2001introduction}, and population-size or survival estimation with capture--recapture models \cite{nichols1992capture}. Adapting the ecological model and the information gathered by expert review to these goals could support a wider range of conservation questions. Multi-species and spatial models could further enable more efficient review across species and locations. Examining the sensitivity to temporal and spatial aggregation would clarify how survey design affects both ecological inference and review efficiency. Another promising direction is to determine when expert effort is better used for fine-tuning classifiers or directly informing downstream models, and how these uses can be combined adaptively. Finally, convergence guarantees could complement
our empirical evidence that \methoda\ converges to the oracle solution.

\clearpage
\section*{Acknowledgements and disclosures}

This work was supported in part by the Schwarzman College of Computing fellowship, NSF fellowship award No. 2313998, the AI and Biodiversity Change (ABC) Global Center, which is funded by the US National Science Foundation under Award No. 2330423 and Natural Sciences and Engineering Research Council of Canada under Award No. 585136, NSF CAREER Award No. 2441060, the MIT Social and Ethical Responsibilities of Computing Initiative, and the MIT-Google program for Computing Innovation and Planetary Health. The authors declare no competing interests.

\bibliography{main}

\newpage
\appendix

\section{Additional details on datasets and covariates}\label{sec:appendix:datasets}\label{sec:appendix:covariates}
We use different preprocessing pipelines for the acoustic forest soundscape and iWildCam camera-trap data before representing each focal species as a single-season occupancy problem. 

\noindent\textbf{Acoustic.}
We use the acoustic forest soundscape dataset \citep{enaa} with Perch v2 classifier outputs \citep{perch_v2}. This dataset is openly available \citep{enaa} and licensed under the \href{https://creativecommons.org/licenses/by/4.0/}{Creative Commons Attribution 4.0 International license}. We treat each recording (typically 10 minutes long) as one occupancy replicate. We keep recordings from the ACAD, MABI, and SIMR subcollections and assign a positive label when the ground truth labels for a file contain the focal species. The classifier score is the maximum target-species Perch v2 logit over an entire recording. Each of the 50 focal bird species is modeled on the same 104 recorder sites and 1,302 reviewable recordings, with at most 28 recordings per site. Across these species and recording labels, there are 8,962 positive detections. The number of per-species detections ranges from 20 to 1,012, with median 116.

The acoustic occupancy model uses three site-level environmental covariates summarized around each recorder: mean tree-canopy cover percentage from MODIS \cite{modis}, median canopy height from GEDI \citep{gedi}, and mean elevation from USGS 3DEP \citep{us20193d}, sampled using Google Earth Engine. The detection covariates are hours since local sunrise, day of year, and binary indicators for the MABI and SIMR subcollections, with ACAD as the reference level. Covariates are median-imputed and z-score normalized.

\noindent\textbf{Camera trap.}
We use iWildCam 2022 \citep{beery2021iwildcam} with SpeciesNet outputs \citep{speciesnet} for our camera-trap analysis. iWildCam 2022 is available at \url{https://lila.science/datasets/iwildcam-2022/} and licensed under the \href{https://cdla.io/permissive-1-0/}{Community Data License Agreement (CDLA)}. We use only the training split, since the test split does not contain ground truth labels. We discard sites that lack geographic locations, and discard images whose capture datetime cannot be parsed. The occupancy site is an iWildCam location and the replicate is a location-day. For each focal species, a location-day is positive when any image from that location-day has the target category, and its classifier score is the maximum SpeciesNet target logit over images from that location-day. We summarize daily effort as the number of unique image sequences on the location-day and use $\log(\max(\text{sequence count}, 1))$ as a detection covariate.

Because sites in iWildCam are distributed across the globe and our goal is to model regional occupancy processes at fine rather than continental scales, we apply species-specific geographic filtering before fitting the occupancy model. We form deterministic connected-component clusters of locations using a 250 km Haversine-distance radius and, for each species, keep only clusters that contain at least one positive location-day. This removes distant negative-only clusters that are unlikely to be informative about the same regional occupancy process as the target detections. The 25 focal camera-trap species have 29 to 98 sites per species (median 42), 692 to 2,365 reviewable location-days per species (median 2,133), and at most 24 to 93 location-days per retained site. Across these species-specific location-day labels there are 44,231 samples and 4,622 positives. The per-species positive count ranges from 41 to 479, with median 151.

The camera-trap occupancy model uses five site-level covariates loosely based on \cite{pease2022}: elevation from SRTM \citep{farr2007shuttle}, forest cover from MODIS \cite{modis}, annual mean temperature and annual mean precipitation from WorldClim \citep{fick2017worldclim}, and mean human population density from GPW 2020 \citep{gpw}, sampled using Google Earth Engine. The camera-trap detection model uses the log daily sequence-count covariate described above as a proxy for sampling effort and trigger sensitivity. Just as for the acoustic data, covariates are median-imputed and z-score normalized.

\begin{figure}
    \centering
    \includegraphics[width=1\linewidth]{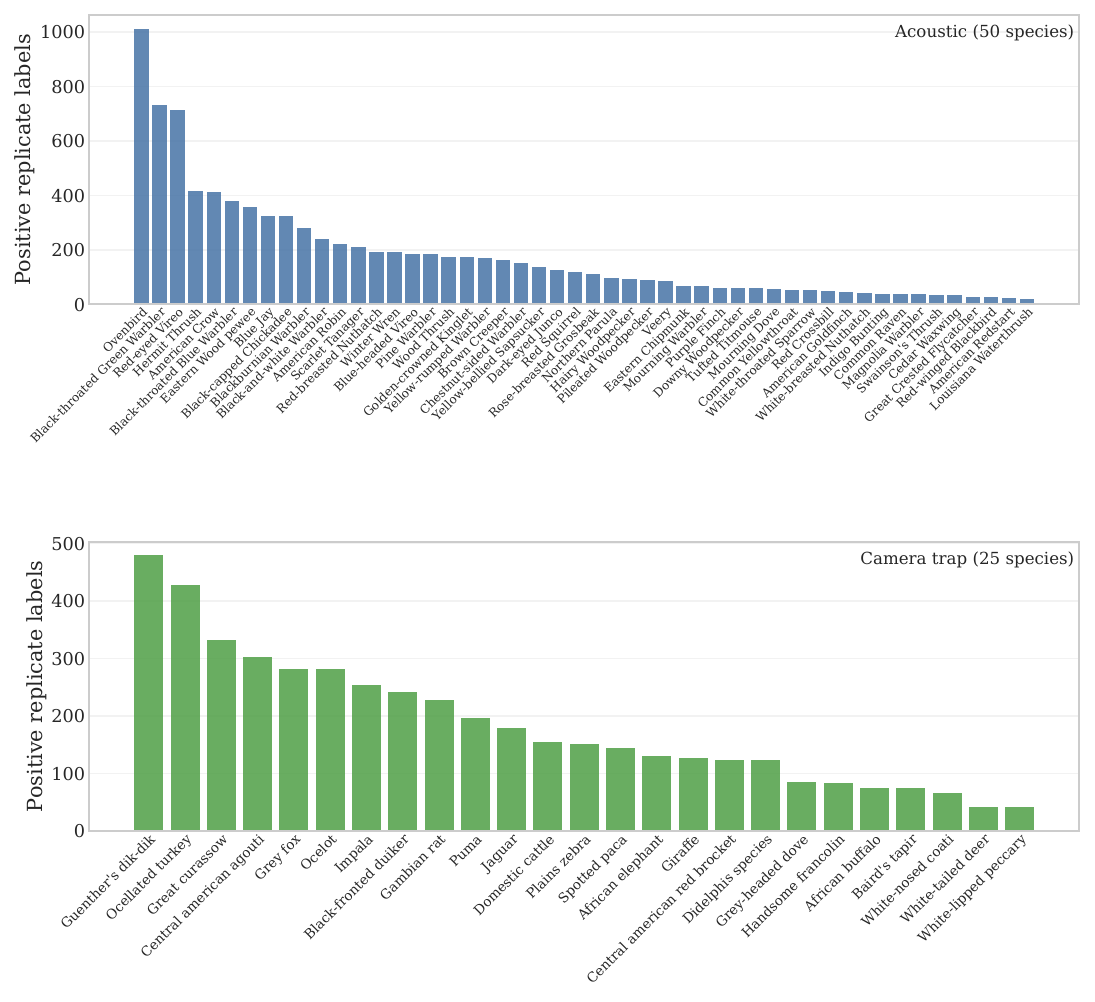}
    \caption{\textbf{Class distribution of focal species used in the real-data experiments.} Bars show the number of positive replicate-level labels for the 50 acoustic species and 25 iWildCam camera-trap species used in the main sweep, sorted from most to least common within each dataset. Acoustic replicates are recordings. Camera-trap replicates are location-days after train-split, geocoding, and positive-geographic-cluster filtering.}
    \label{fig:dataset_class_distribution}
\end{figure}

\section{Experimental details}\label{sec:appendix:computational_cost}\label{sec:appendix:implementation_details}

\subsection{Model fitting}
The Bernoulli and continuous-score occupancy models are implemented using the Biolith library \citep{biolith}. To sample from the posterior, we use the No-U-Turn Sampler (NUTS, \cite{nuts}), a variant of Hamiltonian Monte Carlo \citep{hmc1,hmc2}, implemented using NumPyro \citep{numpyro1,numpyro2}. We use three chains, 500 warmup steps and 500 retained samples per chain. Separate score calibration uses one chain with 100 warmup and 100 retained samples, eight calibration draws for acquisition, and a prevalence prior with mean 0.01 and concentration 100.

\subsection{Computational complexity}
Each active-review iteration has two computationally relevant stages: refitting the current ecological model and ranking candidate reviews. As noted in prior work in reliable statistical computation \cite{misztal2008}, the choice of estimation algorithm can involve substantial tradeoffs between computational efficiency and reliability. In our setting, the refitting stage dominates due to computationally intensive MCMC sampling. For the decoupled continuous-score model, the reported fitting time combines the occupancy-model fit with the separate score-mixture calibration fit. For the original continuous-score model, the score-mixture parameters are fit jointly inside the occupancy-model sampler. Candidate ranking is much cheaper in comparison. Random and max-score selection require only simple ordering or sampling over reviewable replicates, PPU and BALD additionally average over posterior draws, and Target EIG is the most expensive ranking rule because it recomputes a target-posterior entropy approximation for each candidate label outcome. Even for Target EIG, however, acquisition is a few seconds per iteration and remains smaller than posterior fitting (\zcref{tab:computational_cost_iteration_times}).

\begin{table}[t]
\caption{Mean wall-clock time per active-review iteration, with three seeds pooled within each species, averaged over species-experiment trajectories after excluding terminal iterations with no selected reviews. Model fitting is the occupancy-model MCMC time plus score-mixture calibration for the decoupled continuous-score model. Ranking is the time used to score and order candidate reviews. Total includes recorded per-iteration overhead in addition to these two components. Entries show mean $\pm$ 95\% CI half-width, using the normal approximation across species-experiment trajectory means.}
\label{tab:computational_cost_iteration_times}
\centering
\begingroup
\small
\resizebox{\textwidth}{!}{
\begin{tabular}{lllcccc}
\toprule
Dataset & Model & Review method & Species & Model fitting (s, mean $\pm$ 95\% CI) & Ranking (s, mean $\pm$ 95\% CI) & Total (s, mean $\pm$ 95\% CI) \\
\midrule
Acoustic & \multirow{5}{*}{Reviewed-only Bernoulli} & Random & 50 & 8.38 $\pm$ 0.57 & $<0.01$ & 8.64 $\pm$ 0.58 \\
 &  & Max-score & 50 & 8.74 $\pm$ 0.75 & $<0.01$ & 9.00 $\pm$ 0.75 \\
 &  & PPU & 50 & 10.39 $\pm$ 0.92 & $<0.01$ & 10.66 $\pm$ 0.93 \\
 &  & BALD & 50 & 8.78 $\pm$ 0.66 & $<0.01$ & 9.05 $\pm$ 0.67 \\
 &  & Target EIG & 50 & 8.99 $\pm$ 0.47 & 1.84 $\pm$ 0.08 & 11.10 $\pm$ 0.53 \\
\cmidrule(lr){2-7}
 & \multirow{5}{*}{Decoupled continuous-score} & Random & 50 & 10.14 $\pm$ 0.62 & $<0.01$ & 10.84 $\pm$ 0.62 \\
 &  & Max-score & 50 & 10.16 $\pm$ 0.57 & $<0.01$ & 10.85 $\pm$ 0.58 \\
 &  & PPU & 50 & 10.11 $\pm$ 0.63 & $<0.01$ & 10.80 $\pm$ 0.64 \\
 &  & BALD & 50 & 10.05 $\pm$ 0.60 & $<0.01$ & 10.74 $\pm$ 0.60 \\
 &  & Target EIG & 50 & 9.26 $\pm$ 0.56 & 2.29 $\pm$ 0.08 & 12.26 $\pm$ 0.59 \\
\cmidrule(lr){2-7}
 & \multirow{2}{*}{Original continuous-score} & BALD & 50 & 20.72 $\pm$ 1.45 & $<0.01$ & 21.03 $\pm$ 1.46 \\
 &  & Target EIG & 50 & 19.93 $\pm$ 1.43 & 1.98 $\pm$ 0.07 & 22.22 $\pm$ 1.46 \\
\midrule
Camera trap & \multirow{5}{*}{Reviewed-only Bernoulli} & Random & 25 & 6.15 $\pm$ 0.63 & 0.01 $\pm$ 0.00 & 6.40 $\pm$ 0.66 \\
 &  & Max-score & 25 & 6.31 $\pm$ 0.65 & $<0.01$ & 6.50 $\pm$ 0.67 \\
 &  & PPU & 25 & 6.56 $\pm$ 0.71 & $<0.01$ & 6.75 $\pm$ 0.73 \\
 &  & BALD & 25 & 6.58 $\pm$ 0.80 & $<0.01$ & 6.76 $\pm$ 0.82 \\
 &  & Target EIG & 25 & 6.46 $\pm$ 0.73 & 1.65 $\pm$ 0.09 & 8.30 $\pm$ 0.83 \\
\cmidrule(lr){2-7}
 & \multirow{5}{*}{Decoupled continuous-score} & Random & 25 & 8.59 $\pm$ 0.85 & $<0.01$ & 9.27 $\pm$ 0.96 \\
 &  & Max-score & 25 & 8.78 $\pm$ 0.85 & $<0.01$ & 9.46 $\pm$ 0.97 \\
 &  & PPU & 25 & 8.69 $\pm$ 0.88 & $<0.01$ & 9.35 $\pm$ 1.00 \\
 &  & BALD & 25 & 8.71 $\pm$ 0.85 & $<0.01$ & 9.38 $\pm$ 0.96 \\
 &  & Target EIG & 25 & 8.06 $\pm$ 0.83 & 2.28 $\pm$ 0.25 & 10.99 $\pm$ 1.14 \\
\cmidrule(lr){2-7}
 & \multirow{2}{*}{Original continuous-score} & BALD & 25 & 23.91 $\pm$ 3.41 & $<0.01$ & 24.15 $\pm$ 3.44 \\
 &  & Target EIG & 25 & 24.59 $\pm$ 4.77 & 1.69 $\pm$ 0.23 & 26.52 $\pm$ 4.98 \\
\bottomrule
\end{tabular}

}
\endgroup
\end{table}

All experiments were executed without GPUs on a Slurm cluster with one job per species. Each job ran on 16 cores of an x86\_64 AMD EPYC 9554 64-Core CPU with 32 GB of memory.

Across the three seeds, runtime for the all configurations averaged 10.7 hours per acoustic species-seed and 1.5 hours per camera-trap species-seed, totaling approximately 1,718 job-hours. With $P$ concurrent jobs, the estimated wall-clock time to reproduce the current 50 acoustic and 25 camera-trap experiments is
\[
\frac{1{,}718\ \text{job-hours}}{P}.
\]

While this is the approximate time required to reproduce our experiments given the specified hardware, the full research project required more resources in total.

\subsection{Uncertainty quantification}
Confidence intervals (CIs) are pointwise 95\% intervals, without multiple-comparison correction. Normal-approximation CIs use the mean $\pm\,1.96$ times its standard error, with the averaging unit specified in each caption. Unless noted otherwise, seeds are averaged within species before estimating uncertainty. Species-bootstrap CIs use 2,000 resamples of species: percentile CIs use the 2.5th and 97.5th percentiles of bootstrap means, whereas bootstrap-normal CIs use the bootstrap standard error. Posterior credible bounds describe within-fit uncertainty, not uncertainty across species.

\section{Ecological tasks and characteristics of selected data}\label{sec:appendix:task_characterization}

We investigate task-specific differences across methods and compare \methoda\ with random, max-score, BALD, and PPU review under the same decoupled continuous-score model, and with EIG focused separately on site occupancy, occupancy coefficients, or detection coefficients. \zcref{tab:task_specific_performance} shows differences in average oracle agreement over each full review trajectory. Site-occupancy agreement averages $A_\psi$ and $\rho_\psi$, coefficient tasks use $A_\beta$ or $A_\alpha$, and the combined score is $\overline{A}$ (\zcref{sec:experiments:evaluation}). Seeds are averaged within species before resampling species for confidence intervals.

\begin{table}[tbp]
\caption{\textbf{Average agreement over full review trajectories by ecological task.} Entries are paired differences in trajectory-average oracle agreement, in percentage points, shown as mean $\pm$ 95\% CI half-width, using the species-bootstrap normal approximation after averaging paired differences across seeds within species. Positive entries (green) mean \methoda\ performs better while negative entries (red) mean \methoda\ performs worse. Task-specific \methoda\ focuses EIG on the evaluated task.}
\label{tab:task_specific_performance}
\centering
\begingroup
\footnotesize
\definecolor{acornTableGain}{RGB}{44,140,88}
\definecolor{acornTableLoss}{RGB}{190,58,67}
\setlength{\tabcolsep}{2pt}
\begin{tabular}{@{}lrrrrr@{}}
\toprule
 & \multicolumn{5}{c}{\textbf{ACORN vs.}} \\
\cmidrule(lr){2-6}
Evaluation task & Random & Max-score & BALD & PPU & \shortstack{Task-specific\\ACORN} \\
\midrule
\multicolumn{6}{l}{\textbf{Acoustic}} \\
Combined & \cellcolor{acornTableGain!34.9}$+6.50 \pm 1.85$ & \cellcolor{acornTableGain!12.5}$+1.08 \pm 1.02$ & \cellcolor{acornTableGain!20.9}$+3.11 \pm 0.95$ & \cellcolor{acornTableGain!31.4}$+5.65 \pm 1.52$ & -- \\
Site occupancy & \cellcolor{acornTableGain!35.2}$+6.57 \pm 2.10$ & \cellcolor{acornTableGain!15.7}$+1.85 \pm 1.11$ & \cellcolor{acornTableGain!20.3}$+2.97 \pm 0.91$ & \cellcolor{acornTableGain!30.5}$+5.44 \pm 1.52$ & \cellcolor{acornTableGain!8.1}$+0.02 \pm 0.05$ \\
Occupancy coefficients & \cellcolor{acornTableGain!43.1}$+8.48 \pm 3.81$ & \cellcolor{acornTableGain!13.3}$+1.27 \pm 2.02$ & \cellcolor{acornTableGain!32.1}$+5.82 \pm 1.88$ & \cellcolor{acornTableGain!45.0}$+8.93 \pm 2.76$ & \cellcolor{acornTableLoss!8.2}$-0.05 \pm 0.26$ \\
Detection coefficients & \cellcolor{acornTableGain!26.1}$+4.37 \pm 2.90$ & \cellcolor{acornTableLoss!10.7}$-0.64 \pm 1.86$ & \cellcolor{acornTableGain!10.9}$+0.69 \pm 1.12$ & \cellcolor{acornTableGain!19.6}$+2.81 \pm 1.75$ & \cellcolor{acornTableLoss!8.0}$-0.01 \pm 0.95$ \\
\midrule
\multicolumn{6}{l}{\textbf{Camera trap}} \\
Combined & \cellcolor{acornTableGain!29.5}$+5.18 \pm 1.83$ & \cellcolor{acornTableGain!11.7}$+0.90 \pm 0.44$ & \cellcolor{acornTableGain!13.0}$+1.21 \pm 1.36$ & \cellcolor{acornTableGain!13.6}$+1.36 \pm 1.60$ & -- \\
Site occupancy & \cellcolor{acornTableGain!36.5}$+6.88 \pm 3.61$ & \cellcolor{acornTableGain!12.9}$+1.18 \pm 0.69$ & \cellcolor{acornTableGain!11.4}$+0.81 \pm 0.54$ & \cellcolor{acornTableGain!12.1}$+1.00 \pm 0.80$ & \cellcolor{acornTableLoss!8.4}$-0.10 \pm 0.13$ \\
Occupancy coefficients & \cellcolor{acornTableGain!22.0}$+3.38 \pm 2.31$ & \cellcolor{acornTableGain!11.5}$+0.84 \pm 0.54$ & \cellcolor{acornTableGain!12.8}$+1.16 \pm 1.41$ & \cellcolor{acornTableGain!13.9}$+1.42 \pm 1.70$ & \cellcolor{acornTableGain!8.5}$+0.13 \pm 0.23$ \\
Detection coefficients & \cellcolor{acornTableGain!22.8}$+3.58 \pm 3.02$ & \cellcolor{acornTableGain!9.7}$+0.40 \pm 0.83$ & \cellcolor{acornTableGain!16.5}$+2.06 \pm 3.32$ & \cellcolor{acornTableGain!16.5}$+2.04 \pm 3.36$ & \cellcolor{acornTableGain!15.0}$+1.68 \pm 2.68$ \\
\bottomrule
\end{tabular}

\endgroup
\end{table}

Over full trajectories, \methoda\ improves combined agreement over max-score by 1.08 percentage points on acoustics and 0.90 on camera traps. Averaging agreement over the first 100 reviews instead gives larger gains of 4.79 points (95\% species-bootstrap percentile CI: $[2.32,7.36]$) and 2.44 points ($[0.74,4.03]$), respectively, computed as the area under the agreement curve divided by the review budget. The largest gains over random review across full trajectories are for acoustic occupancy coefficients (8.48 points) and camera-trap site occupancy (6.88 points). Gains are not uniform: acoustic detection-coefficient agreement is 0.64 points lower than max-score on average. Interestingly, focusing EIG on the evaluated task changes mean agreement by less than 0.14 points, except for camera-trap detection-only targeting, which falls 1.68 points behind combined targeting. Thus, task specialization does not consistently improve performance. We hypothesize this is because our occupancy modeling tasks are inherently coupled: errors in the estimation of the occupancy process force the detection process to compensate, and vice-versa.

\begin{table}[tbp]
\caption{\textbf{Characteristics of selected replicates over the first 100 reviews.} Entries average all selected replicates over the first 100 reviews. Bold marks the highest displayed value in each column within each dataset. \emph{New site} is the percentage of sites with no prior reviewed label. \emph{Score rank} and \emph{covariate extremeness} are mean percentiles within the current unreviewed candidate pool. Extremeness is squared Mahalanobis distance in the occupancy or detection covariates. \emph{Label confidence} is $100\max(p,1-p)$ for the ecological model's posterior predictive label probability $p$.}
\label{tab:selected_sample_characteristics}
\centering
\begingroup
\footnotesize
\setlength{\tabcolsep}{3pt}
\begin{tabular}{lrrrrr}
\toprule
Review policy & \shortstack{New\\site} & \shortstack{Score\\rank} & \shortstack{Label\\confidence} & \shortstack{Occupancy\\covariate\\extremeness} & \shortstack{Detection\\covariate\\extremeness} \\
\midrule
\multicolumn{6}{l}{\textbf{Acoustic}} \\
Random & 56.0 & 52.6 & \textbf{96.3} & 52.8 & 48.2 \\
Max-score & 41.1 & \textbf{99.8} & 89.5 & 47.2 & 47.8 \\
BALD & 48.8 & 87.1 & 82.0 & 50.2 & 54.0 \\
PPU & 38.2 & 84.0 & 70.7 & 49.3 & 47.2 \\
ACORN (combined) & 57.7 & 87.7 & 88.9 & \textbf{53.7} & 58.5 \\
EIG: site occupancy & 57.9 & 87.5 & 89.1 & \textbf{53.7} & 58.4 \\
EIG: occupancy coefficients & \textbf{58.0} & 87.6 & 89.0 & \textbf{53.7} & 58.5 \\
EIG: detection coefficients & 45.9 & 87.4 & 82.6 & 49.9 & \textbf{59.8} \\
\midrule
\multicolumn{6}{l}{\textbf{Camera trap}} \\
Random & 51.4 & 48.9 & \textbf{98.9} & \textbf{50.5} & 51.5 \\
Max-score & 48.3 & \textbf{99.2} & 96.9 & 47.7 & 52.2 \\
BALD & 57.1 & 90.7 & 89.9 & 48.9 & 56.0 \\
PPU & 47.7 & 86.4 & 86.0 & 49.2 & 53.1 \\
ACORN (combined) & 64.1 & 91.1 & 91.9 & 49.6 & 55.4 \\
EIG: site occupancy & \textbf{66.5} & 90.3 & 92.2 & 49.8 & 55.4 \\
EIG: occupancy coefficients & 65.9 & 90.8 & 92.1 & 49.9 & 55.7 \\
EIG: detection coefficients & 56.4 & 91.1 & 90.2 & 48.9 & \textbf{58.5} \\
\bottomrule
\end{tabular}

\endgroup
\end{table}

We furthermore investigate characteristics of the data that is prioritized by each method for the first 100 reviewed replicates in \zcref{tab:selected_sample_characteristics}. Combined-target \methoda\ does not simply follow classifier-score order: mean current-pool score percentiles are 87.7 for acoustics and 91.1 for camera traps, compared with 99.8 and 99.2 for max-score. Detection-focused EIG has the highest mean visit-covariate extremeness percentile in both datasets (59.8 vs. 58.5 for combined targeting in acoustics, 58.5 vs. 55.4 in camera traps). Site-occupancy targeting selects more reviews from previously unreviewed sites than detection targeting (57.9\% vs. 45.9\% for acoustics, 66.5\% vs. 56.4\% for camera traps).

Different targets can nevertheless select distinct review sets: after 100 acoustic reviews, the mean Jaccard overlap (intersection divided by union) is 0.43 between site-occupancy and detection-coefficient targeting, versus 0.95 between site-occupancy and occupancy-coefficient targeting. Similar ecological agreement can therefore arise from different sets of reviewed observations.

Dataset characteristics also matter. For example, acoustic gains in averaged oracle agreement over max-score correlate with positive prevalence, site occupancy, and empirical detection probability (Spearman $\rho=0.45$, $0.46$, and $0.39$), and decrease when positives are concentrated at fewer sites ($\rho=-0.39$). Mean classifier average precision is higher for camera traps (0.95 versus 0.74 for acoustics). These post-hoc associations do not establish why gains differ between datasets.

\clearpage
\section{Sensitivity to review batch size}\label{sec:appendix:batch_size}

The batch size $m$ controls how many expert reviews are collected before refitting the model and updating the review priorities. We use $m=5$ for acoustic data and $m=25$ for camera traps. To characterize the tradeoff between computational cost and responsiveness to new labels, we fit the decoupled continuous-score model with four policies (Target EIG, BALD, random, and max-score), and a budget of 100 reviews. We evaluate $m \in \{1,2,5,10,20\}$ for acoustic data and $m \in \{5,10,25,50,100\}$ for camera traps.

To assess sensitivity without relying on a fully reviewed oracle, we compare each trajectory with the finest cadence for the same species, policy, and seed ($m_{\min}=1$ acoustic, $m_{\min}=5$ camera trap). At a shared review count $R$, we measure
\begin{equation}\label{eq:batch_size_sensitivity}
\Delta_\psi^{(m)}(R)=\frac{1}{|S|}\sum_{i\in S}\left|\bar\psi_i^{(m)}(R)-\bar\psi_i^{(m_{\min})}(R)\right|.
\end{equation}
We also compare coefficient conclusions using the definitions in Eq.~\ref{eq:cov}, with the finest cadence replacing the oracle.

\begin{figure}[tbp]
    \centering
    \begin{subfigure}{\linewidth}
        \centering
        \includegraphics[width=\linewidth]{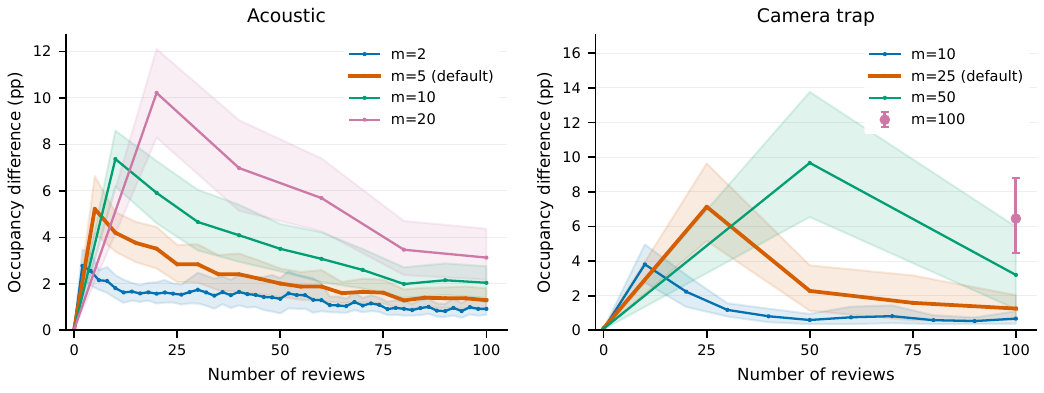}
        \caption{Occupancy differences over the review trajectory, relative to the finest cadence ($m=1$ acoustic, $m=5$ camera trap).}
        \label{fig:batch_size_ablation:trajectories}
    \end{subfigure}
    \par\medskip
    \begin{subfigure}{\linewidth}
        \centering
        \includegraphics[width=\linewidth]{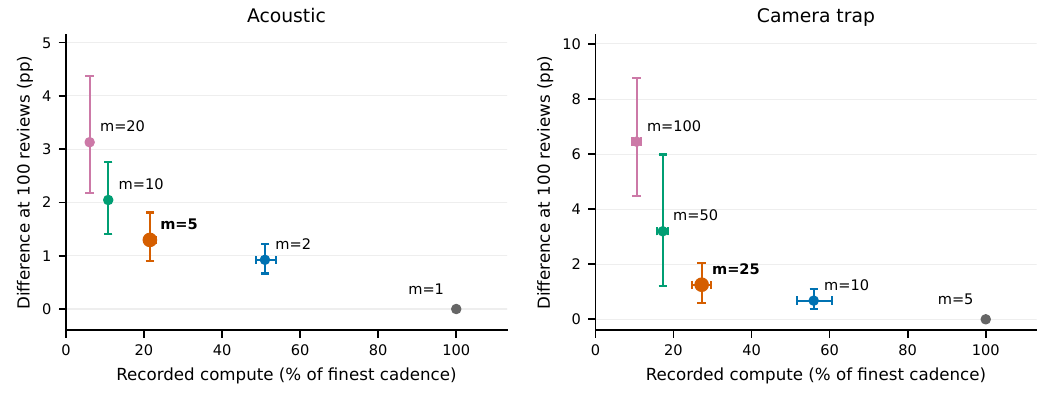}
        \caption{Recorded compute vs. occupancy differences at 100 reviews. Bold labels mark our defaults ($m=5$ acoustic, $m=25$ camera trap).}
        \label{fig:batch_size_ablation:runtime}
    \end{subfigure}
    \caption{\textbf{Larger review batches reduce computation but change early posterior trajectories.} All panels show Target EIG, comparing each $m$ with the finest cadence instead of the oracle. Occupancy differences are mean absolute differences in site occupancy probabilities, in percentage points (pp). Bands and error bars show pointwise 95\% species-bootstrap percentile CIs after averaging seeds within species.}
    \label{fig:batch_size_ablation}
\end{figure}

\zcref{fig:batch_size_ablation,tab:batch_size_ablation} show the results for Target EIG. Larger batches have an early impact: after 20 acoustic reviews, mean occupancy differences from the finest cadence are 3.51 percentage points for $m=5$ and 10.21 for $m=20$. After 50 camera-trap reviews, they are 2.27 points for $m=25$ and 9.67 for $m=50$. By 100 reviews, the differences at the defaults decrease to 1.30 and 1.26 points, respectively. The defaults require 21.5\% and 27.2\% of the recorded computation of the finest cadence. Thus, they offer substantial computational savings while retaining more frequent feedback than the coarser alternatives. Interestingly, larger batches do not uniformly increase disagreement between review policies. Relative to $m=5$, using $m=25$ increases camera-trap Target EIG--BALD occupancy disagreement at 50 reviews by 1.1 percentage points (95\% species-bootstrap percentile CI: $[0.2,2.1]$), but using $m=50$ reduces it by 1.9 points ($[-2.7,-1.2]$).

\begin{table}[t]
\caption{\textbf{Batch-size sensitivity and computational cost for Target EIG.} Comparisons use the finest cadence instead of the oracle ($m=1$ acoustic, $m=5$ camera trap). Occupancy differences are mean absolute site-probability differences in percentage points (pp). Early comparisons use 20 acoustic or 50 camera-trap reviews. Coefficient agreement at 100 reviews uses the positive, negative, or uncertain conclusions defined in Eq.~\ref{eq:cov}. Bold batch sizes are the default.}
\label{tab:batch_size_ablation}
\centering
\begingroup
\small
\setlength{\tabcolsep}{4pt}
\begin{tabular}{lrrrrrrr}
\toprule
Dataset & $m$ & Fits & Time & \multicolumn{2}{c}{Occupancy difference (pp)} & \multicolumn{2}{c}{Coeff. agreement (\%)} \\
 & & & (\%) & Early & At 100 & Occupancy & Detection \\
\midrule
\multirow{5}{*}{Acoustic} & 1 & 101 & 100.0 & 0.00 & 0.00 & 100.0 & 100.0 \\
 & 2 & 51 & 51.0 & 1.58 & 0.92 & 98.2 & 96.4 \\
 & \textbf{5} & 21 & 21.5 & 3.51 & 1.30 & 98.0 & 96.5 \\
 & 10 & 11 & 10.8 & 5.91 & 2.05 & 96.3 & 94.8 \\
 & 20 & 6 & 6.1 & 10.21 & 3.13 & 94.0 & 92.1 \\
\midrule
\multirow{5}{*}{Camera trap} & 5 & 21 & 100.0 & 0.00 & 0.00 & 100.0 & 100.0 \\
 & 10 & 11 & 55.9 & 0.60 & 0.68 & 99.3 & 100.0 \\
 & \textbf{25} & 5 & 27.2 & 2.27 & 1.26 & 98.7 & 98.7 \\
 & 50 & 3 & 17.2 & 9.67 & 3.20 & 97.6 & 98.7 \\
 & 100 & 2 & 10.5 & -- & 6.45 & 95.1 & 98.0 \\
\bottomrule
\end{tabular}

\endgroup
\end{table}

\clearpage
\section{Sensitivity to review granularity}\label{sec:appendix:review_granularity}

In the main experiments, we simulate expert review across all samples in a selected replicate (an acoustic recording or a camera trap location-day). However, when expert time is scarce and the classifier scores true observations highly, this might be wasteful.

We therefore compare default \methoda\ using full-replicate labels with labels from only the highest-scoring five-second acoustic window, camera image, or camera sequence (the sequence containing the highest-scoring image within that location-day). Crucially, if the highest-scoring sample is a negative, the entire replicate is treated as negative, potentially missing true detections that are lower-scoring. We compare the impact of this highest-scoring review strategy using the same full-replicate oracle described in \zcref{sec:methods}.

\begin{figure}[tbp]
    \centering
    \includegraphics[width=\linewidth]{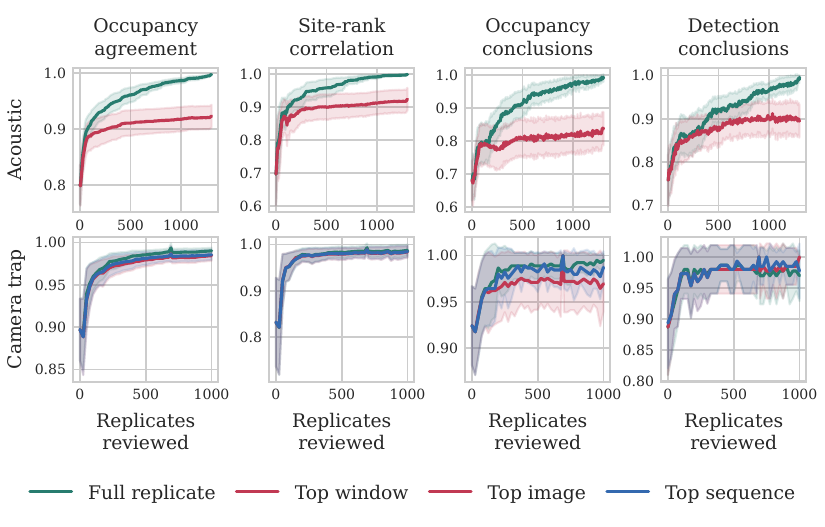}
    \caption{\textbf{Highest-scoring review saves time, but can impact oracle agreement.} We compare different review granularities with the default full-replicate oracle. The x-axis shows the total number of reviews, while the y-axis shows agreement with the oracle (higher is better). Shaded bands show pointwise 95\% normal-approximation CIs across species after averaging seeds within species.}
    \label{fig:single_example_review}
\end{figure}

\zcref{fig:single_example_review} shows the resulting review trajectories. Early camera-trap agreement is nearly unchanged, although final occupancy-coefficient agreement is 2.2 percentage points lower for top-image review. For acoustics, mean agreement after all 1,302 reviews is 0.895 with top windows versus 0.996 with full recordings. Therefore, in our experiments, the highest-scoring review strategy is more promising for camera traps than for acoustics, likely because classifier performance is higher on camera traps (\zcref{sec:appendix:classifier}).

At 100 reviews, highest-scoring review reduces the length of reviewed audio by approximately 120-fold and camera-trap image counts by 64-fold (top image) or nine-fold (top sequence).

\clearpage
\section{Score distribution heterogeneity}\label{sec:appendix:score_heterogeneity}

\begin{figure}
     \centering
     \begin{subfigure}[b]{0.49\textwidth}
         \centering
         \includegraphics[width=\linewidth]{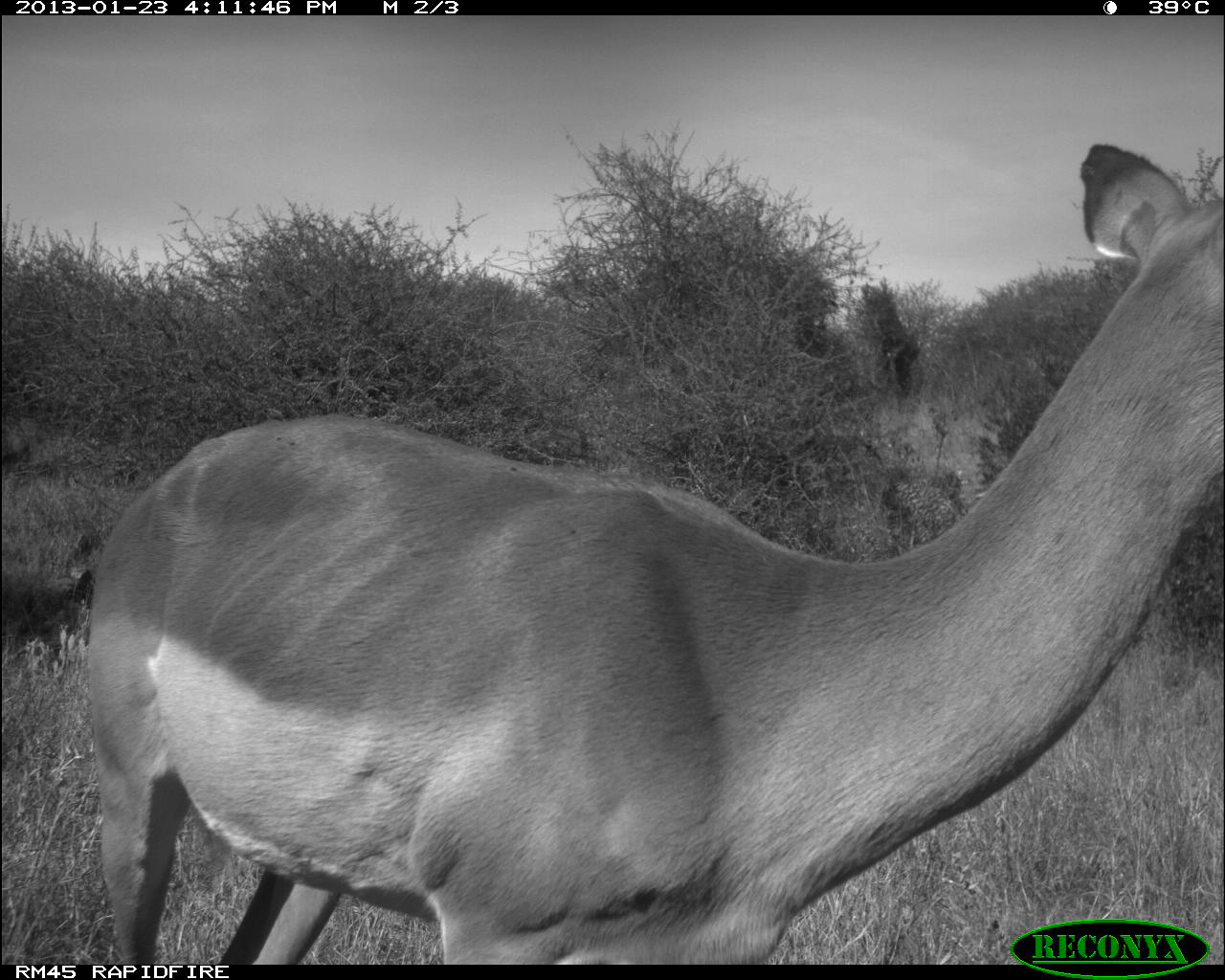}
         \caption{Top-scoring nominal negative sample for Impala (SpeciesNet logit for Impala: $14.09$, Ground-truth label: Plains zebra, Image ID: 9214d000-21bc-11ea-a13a-137349068a90)}
     \end{subfigure}
     \hfill
     \begin{subfigure}[b]{0.49\textwidth}
         \centering
         \includegraphics[width=\linewidth]{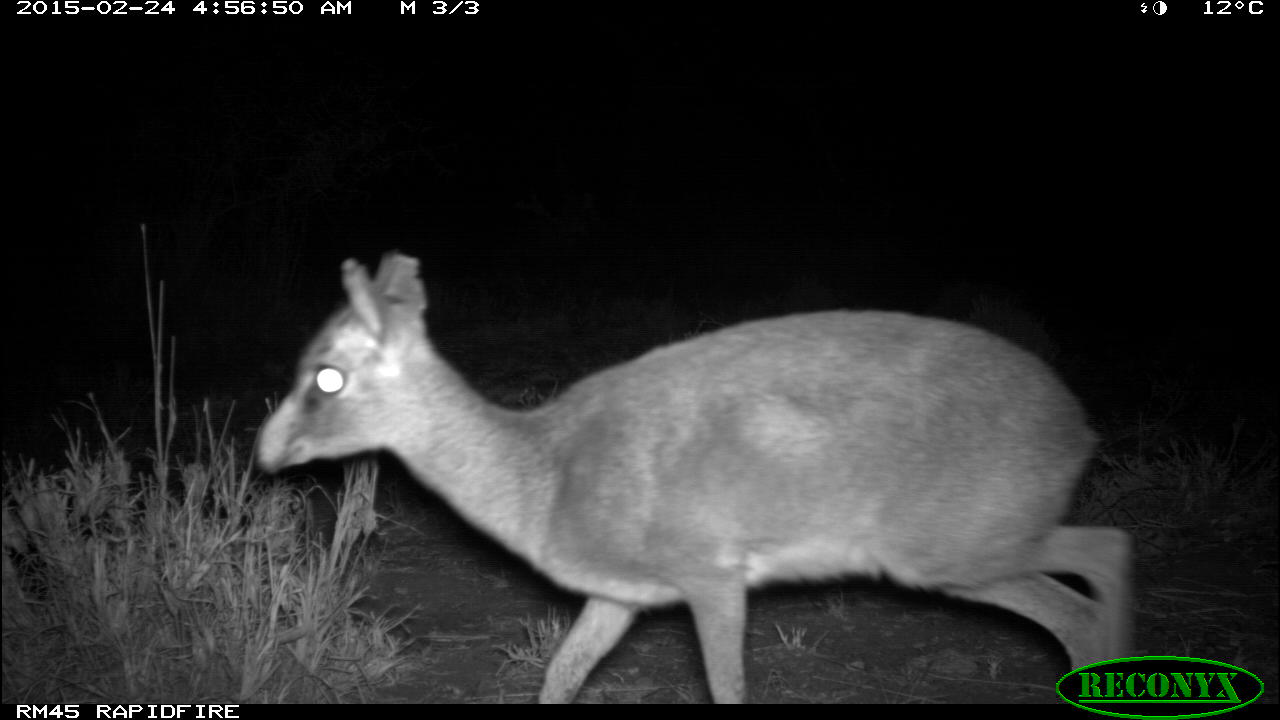}
         \caption{Top-scoring nominal negative sample for Günther's dik-dik (SpeciesNet logit for Günther's dik-dik: $17.08$, Ground-truth label: Grévy's zebra, Image ID: 8fbd0584-21bc-11ea-a13a-137349068a90)}
     \end{subfigure}
    \caption{Example camera trap images corresponding to obvious ground-truth label errors surfaced by highly scoring nominal negatives across two species.}
    \label{fig:mislabeled_examples}
\end{figure}

The continuous-score model assumes that classifier scores are generated from a mixture of normal score distributions that are shared between all sites. In practice, this approximation is not equally good for all species. In post-hoc experiments on both acoustic and iWildCam camera-trap data, several species were well described by a simple two-component score model, but others showed clear score-distribution heterogeneity: scores from non-detection replicates at occupied sites were not always distributed like scores from truly unoccupied sites. Some high-confidence ``no-detection'' days in iWildCam then appeared on manual inspection to contain real detections missed by the annotation process (see \zcref{fig:mislabeled_examples}), creating a high-score tail among nominal negatives (see \zcref{tab:score_heterogeneity_examples} for illustration). Another contributing factor may be classifier bias caused by spurious correlations: visual or acoustic backgrounds that are more likely to contain the focal species might be scored higher, even if the focal species is not present.

\begin{table}[]
\caption{\textbf{Nominal negative score distributions can differ between occupied and likely unoccupied sites.} Values are mean raw logits. Occupied sites have at least one positive observation ($\exists j: f_{ij}=1$), whereas likely unoccupied sites are defined as sites without positive observations ($f_{ij}=0 \forall j$). The table shows species with large positive gaps between negative replicates at occupied and likely unoccupied sites within each dataset.}
\label{tab:score_heterogeneity_examples}
\begin{tabularx}{\textwidth}{llXXX}
\toprule
Dataset & Species & Positive Mean & Negative Mean at Occupied Sites & Negative Mean at Likely Unoccupied Sites \\ \midrule
\multirow{3}{2em}{Acoustic} & Scarlet Tanager        & 10.96 & 8.36 & 7.26 \\
 & Ovenbird               & 12.48 & 8.25 & 7.23 \\
 & Rose-breasted Grosbeak & 10.00 & 7.31 & 6.39 \\
 \midrule
\multirow{3}{2em}{Camera trap} & Impala                 & 14.00 & 6.74 & 5.00 \\
 & Puma                   & 18.92 & 4.46 & 2.80 \\
 & Jaguar                 & 17.25 & 4.53 & 2.94 \\ \bottomrule
\end{tabularx}
\end{table}

This heterogeneity makes the original joint continuous-score formulation more fragile than the calibrated variant used in our main experiments. In the original continuous-score occupancy model formulation \cite{rhinehart2022continuous}, the score-mixture parameters and ecological parameters are estimated together, so misspecification of the score likelihood can feed back into occupancy and detection estimates. The sampler can also encounter weakly identified mixture geometries, because shifts in the score distributions, detection probability, and occupancy probability can explain similar patterns in the observed scores. Decoupling blocks this feedback by fitting score calibration outside the occupancy sampler and passing in fixed log Bayes factors. Moreover, once a replicate is reviewed, the reviewed binary label replaces the score contribution in the occupancy likelihood, preventing a small number of reviewed high-score outliers from directly distorting the ecological posterior through the score model.

We experimented with several alternative joint models to address this misspecification directly, including a separate false-score distribution for occupied sites, noisy-annotation models, and annotation-primary variants with a missed-label score component. These variants sometimes improved agreement with the oracle in particular review regimes, especially at higher review counts, but no single alternative was robust across species and review budgets. The more flexible models often introduced weakly identified mixture components and convergence problems, while fixed-tail annotation-primary models required species- and review-count-specific tuning. We therefore use the calibrated continuous-score model for the main analysis and treat models that allow richer score heterogeneity, particularly for difficult-to-label species, as an important direction for future work.

\section{Covariate shift}\label{sec:appendix:covariate_shift}
New surveys may differ in habitat composition, potentially affecting both occupancy and classifier scores. We ask whether decoupling makes ecological inference more robust to these changes. We compare joint and decoupled refits under covariate shift alone and when classifier scores also change with habitat.

We compared both models without reviewed labels in 30 paired simulations (100 sites, ten visits per site). We shifted $X\sim\mathcal{N}(0,1)$ to $\mathcal{N}(1.5,1)$ while keeping occupancy and detection relationships fixed. A second condition added $0.75X$ to both class-conditional score means, introducing habitat-dependent score bias.

Covariate shift alone shows no clear difference in site-occupancy RMSE (decoupled minus joint: $-0.007$, 95\% CI: $[-0.023,0.009]$). When scores also shift, decoupling improves occupancy-slope recovery (absolute-error difference $-0.204$, CI: $[-0.312,-0.096]$) but worsens site predictions (RMSE difference $0.035$, CI: $[0.011,0.058]$). In these simulations, decoupling can therefore help improve coefficient recovery without improving predictions. We note that these simulations are necessarily artificial, and that we leave investigation into more realistic shifts for future work.

\section{Impact of classifier quality}\label{sec:appendix:classifier}
\begin{figure}
    \centering
    \includegraphics[width=1\linewidth]{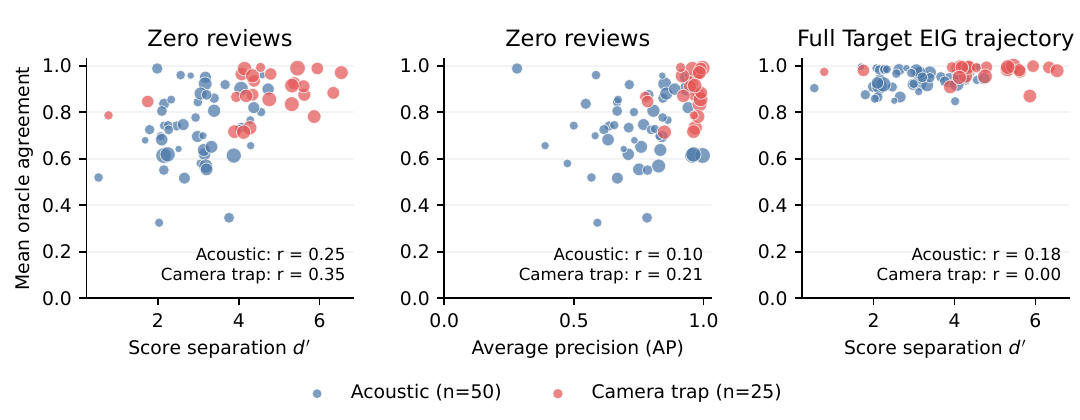}
    \caption{\textbf{Species-level classifier metrics only partially explain downstream ecological agreement.} Each point is one species. Agreement is the mean of the four oracle-agreement metrics (\zcref{sec:experiments:evaluation}). Left: zero-review agreement versus score separation $d'$. Middle: zero-review agreement versus average precision (AP). Right: Target EIG reviewed-trajectory agreement versus $d'$. Point size is proportional to empirical site occupancy prevalence.}
    \label{fig:classifier_quality_agreement}
\end{figure}

We summarize classifier quality in two ways. First, we use score separation $d'$ as the distance between the positive- and negative-score calibration means in units of standard-deviation. Second, we use average precision (AP), which is the mean precision achieved as true positive replicates are swept from high to low classifier score. Species with larger $d'$ tended to start closer to the oracle at zero reviews, but the association was only moderate (\zcref{fig:classifier_quality_agreement}). AP was an even weaker predictor of zero-review agreement. After review using the Target EIG policy (\zcref{sec:methods:policies}), agreement was high regardless of classifier quality. This indicates that improvements in classification accuracy translate into improvements in low-review regimes, but review remains valuable across varying classifier quality.

\section{Application to custom datasets}\label{sec:appendix:practical}

\methoda\ requires repeated samples across sites, suitable site- and replicate-level covariates, expert review capacity, and a classifier covering the focal species. \methoda\ models one species at a time, with occupancy assumed constant within the survey season. Multi-species, multi-season and spatial random-effects models are promising future work.

As discussed in \zcref{sec:appendix:classifier}, imperfect classifiers can still supply useful evidence, but regional differences in backgrounds or sensors may shift the class-conditional score distributions. While decoupling reduces feedback from score-model misspecification, calibration diagnostics and sensitivity analyses remain critical when applying \methoda\ to new datasets. Finally, stopping thresholds should be selected according to the cost of expert review and the desired stability of the ecological conclusions, rather than being interpreted as universal convergence criteria.

\section{Parameter trajectories}\label{sec:appendix:additional_results}

\begin{figure}
    \centering
    \includegraphics[width=1\linewidth]{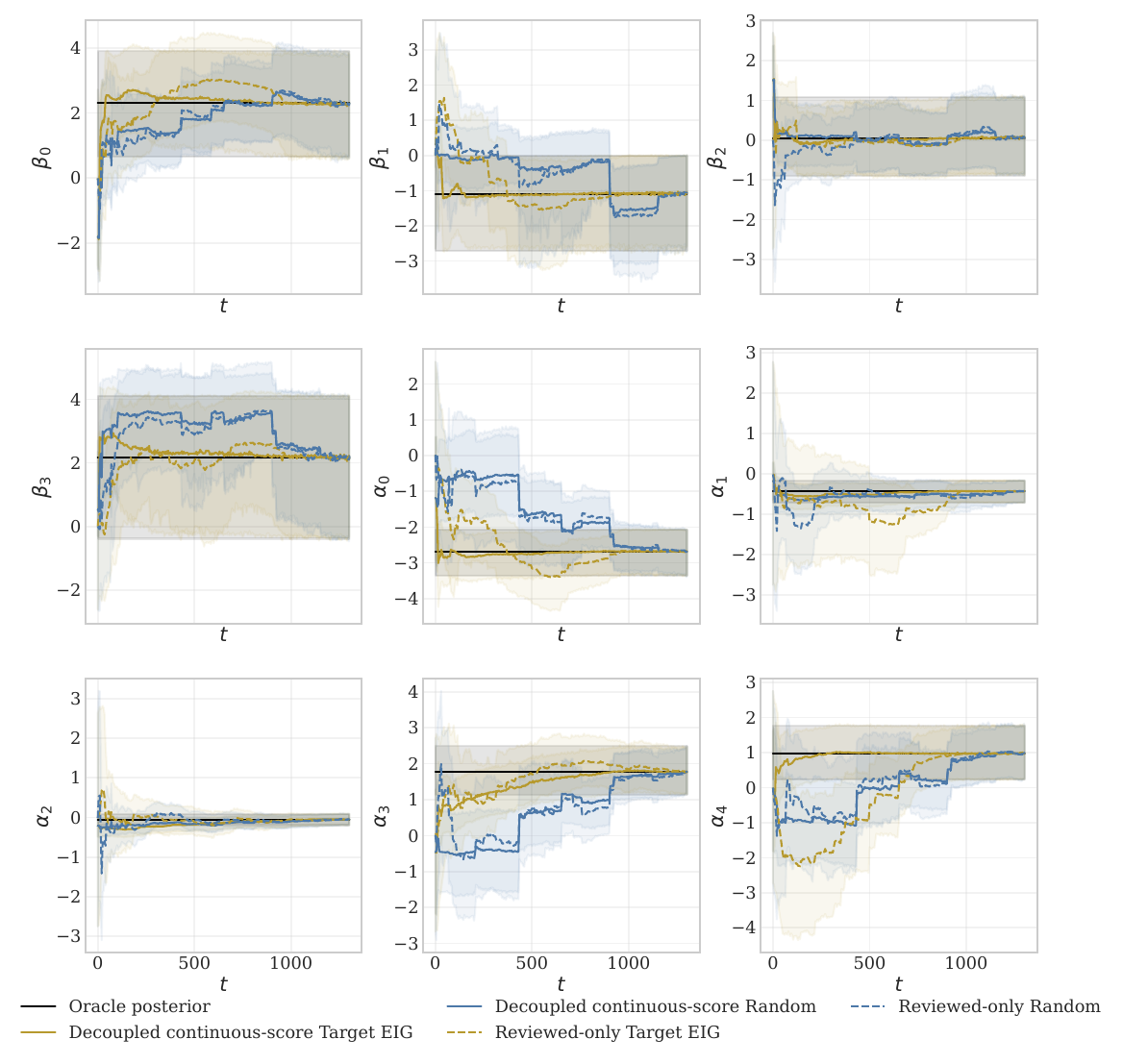}
    \caption{Parameter estimates for the Scarlet Tanager. Shaded bands show within-seed 90\% posterior credible bounds (5th and 95th percentiles), averaged across three seeds.}
    \label{fig:species_example_acoustic}
\end{figure}

\begin{figure}
    \centering
    \includegraphics[width=1\linewidth]{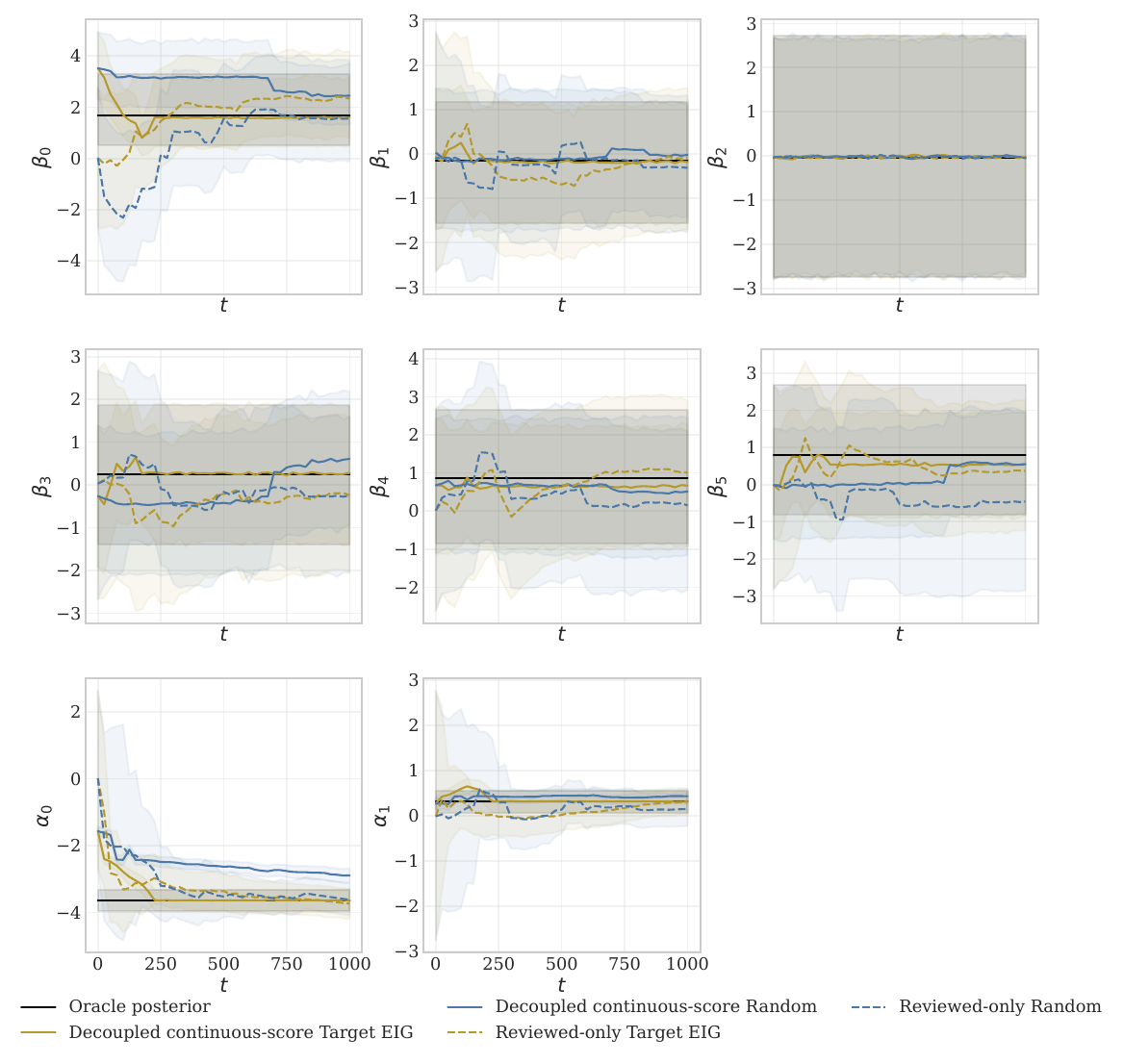}
    \caption{Parameter estimates for the White-tailed deer. Shaded bands show within-seed 90\% posterior credible bounds (5th and 95th percentiles), averaged across three seeds.}
    \label{fig:species_example_camtrap}
\end{figure}

To illustrate qualitatively how parameter estimates converge to the oracle estimates, we visualize species-specific parameter trajectories in \zcref{fig:species_example_acoustic,fig:species_example_camtrap}.

\clearpage
\newpage

\end{document}